\documentclass[svgnames]{new_tlp}

\usepackage{amsmath}
\usepackage{amssymb}
\usepackage{booktabs}
\usepackage{xspace}
\usepackage[
textwidth=0.25\textwidth,textsize=footnotesize]{todonotes}
\newcommand{\mclcomm}[1]{\todo[bordercolor=yellow,linecolor=yellow]{\textsf{\scriptsize MCL: #1}}\xspace}
\newcommand{\mclcommin}[1]{\todo[inline,bordercolor=yellow,linecolor=yellow]{\textsf{\scriptsize MCL: #1}}}

\newcommand{\comm}[1]{\todo[backgroundcolor=white,linecolor=yellow]{\textsf{\scriptsize #1}}\xspace} 
\newcommand{\commin}[1]{\todo[inline,backgroundcolor=white,linecolor=yellow]{\textsf{\scriptsize #1}}\xspace}
\newcommand\lrul[2]{\ensuremath{#1\ \leftarrow \ #2}}

\usepackage{multicol} 
\usepackage{balance}
\usepackage{verbatim}

\let\oldc\, 
\renewcommand{\,}{\oldc \allowbreak}
\let\oldland\land
\renewcommand{\land}{\oldland \allowbreak}
 
\let\oldlor\lor
\renewcommand{\lor}{\oldlor \allowbreak} 

\let\oldlongrightarrow\longrightarrow
\renewcommand{\longrightarrow}{\oldlongrightarrow \allowbreak}

\definecolor{PrologPredicate}{RGB}{0,0,200}
\definecolor{PrologVar}      {RGB}{145,032,039}
\definecolor{PrologComment}  {RGB}{169,082,044}
\definecolor{PrologOther}    {rgb}{0.1,0.1,0.1}
\definecolor{PrologString}   {rgb}{0.2,0.2,0.2}

\newcommand{\ap}{\color{PrologOther}\text{(}}
\newcommand{\cp}{\color{PrologOther}\text{)}}

\usepackage{listings}
\usepackage{textcomp}
\lstdefinestyle{tree}
{
  basicstyle = \small\ttfamily\color{PrologPredicate},
  basewidth = 0.5em,
  moredelim = {[s][\color{PrologString}]{ \{}{\} }},
  moredelim = {*[s][{\color{PrologVar}}]{(}{)}},
  literate     =
  {,}{{,\;}}1
  {\\=}{{\ \char"5C=\ }}3
  {\#<}{{\ \#<\ }}4
  {\#>}{{\ \#>\ }}4
  {\#=}{{\ \#=\ }}3
  {\#<>}{{\ \#<>\ }}5
  {\#=<}{{\ \#=<\ }}5
  {\#>=}{{\ \#>=\ }}5
  {│}{{$|\ $}}1
}
\lstdefinestyle{MySCASP}
{
  keywords = {},
  upquote = true,
  basicstyle = \relsize{-0.5}\ttfamily\color{PrologPredicate},
  basewidth = 0.48em,
  moredelim = {**[is][\color{PrologComment}]{`}{`}},
  moredelim = {*[s][\color{black!40!PrologPredicate}]{\#pred}{.}},
  moredelim = {*[s][\color{black!40!PrologPredicate}]{\#show}{.}},
  moredelim = {*[s][\color{black!40!PrologPredicate}]{\#hide}{.}},
  moredelim = {*[s][\color{PrologVar}]{(}{)}},
  moredelim = {*[s][\color{PrologString}]{'}{'}},
  moredelim = {*[s][\color{PrologOther}]{:-}{.}},
  moredelim = {*[s][\color{red}]{/*}{*/}},
  commentstyle = \mdseries\color{PrologComment},
  morecomment=[l]\%,
  morecomment=[s]{/*}{*/},
  literate     =
  {|}{{$\mid$}}1
  {\ │}{{$\mid$}}1
  {[}{{\color{PrologOther}\small[}}1
  {]}{{\color{PrologOther}\small]}}1
  {\\$}{{\$}}1
  {&(}{{\color{PrologOther}(}}1
  {&)}{{\color{PrologOther})}}1
  {&.}{{.}}0
  {\\=}{{\char"5C=}}2
  {\\$}{{\$}}1,
}
\lstdefinestyle{myASP}
{
   keywords = {},
  upquote = true,
  basicstyle = \small\ttfamily\color{PrologPredicate},
  basewidth = 0.52em,
  moredelim = {*[s][\color{PrologOther}]{:-}{.}},
  moredelim = {*[s][\color{PrologOther}]{\{}{\}}},
  moredelim = {*[s][\color{PrologOther}]{'}{'}},
  moredelim = {*[s][\color{PrologVar}]{(}{)}}, 
  commentstyle = \color{PrologComment},
  morecomment=[l]\%,
  literate     =
  {+}{{+}}2
  {.=.}{{\color{PrologOther}\#=}}3
  {.<.}{{\color{PrologOther}\#<}}3
  {.>.}{{\color{PrologOther}\#>}}3
  {.=<.}{{\color{PrologOther}\#=<}}4
  {.>=.}{{\color{PrologOther}\#>=}}4
  {.\\=.}{{\color{PrologOther}{\charc"5C}=}}3
  {\\=}{{\color{PrologOther}\char"5C=}}3
  {..}{..}2
  {,}{{\color{PrologOther}\footnotesize,}}1
}
\lstdefinestyle{MyInline}
{
  basicstyle = \ttfamily\color{PrologOther},
  breaklines = true,
  breakatwhitespace = true,
  literate =
  {\{}{{\color{PrologOther}\{}}1
  {\}}{{\color{PrologOther}\}}}1
  {:-}{{\color{PrologOther}:\vspace{-.5em}-}}4
  {?-}{{\color{PrologOther}?\vspace{-.5em}-}}4
  {.=.}{{\color{PrologOther}\#=}}3
  {.<.}{{\color{PrologOther}\#<}}3
  {.>.}{{\color{PrologOther}\#>}}3
  {.=<.}{{\color{PrologOther}\#=<}}4
  {.>=.}{{\color{PrologOther}\#>=}}4
  {.\\=.}{{\color{PrologOther}{\charc"5C}=}}3
  {\\=}{{\color{PrologOther}\char"5C=}}3
  {..}{..}2
}
\newcommand{\code}{\lstinline[style=MyInline]}
\usepackage{etoolbox}
\makeatletter
\patchcmd{\lsthk@SelectCharTable}{%
  \lst@ifbreaklines\lst@Def{`)}{\lst@breakProcessOther)}\fi
}{}{}{}
\makeatother

\usepackage{enumitem}
\setitemize{itemsep=1ex}
\setitemize[enumerate]{itemsep=1ex}
\usepackage[justification=centering, skip=5pt]{caption}
\usepackage[skip=5pt]{subcaption}
\newtheorem{example}{Example}
\newtheorem{definition}{Definition}

\usepackage{natbib}
\renewcommand{\cite}{\citep}

\usepackage{url}
\usepackage[hidelinks]{hyperref}
\usepackage{relsize}
\renewcommand*{\UrlFont}{\ttfamily\small\relax}

\title[Modeling and Reasoning in Event Calculus using s(CASP)]{%
  Modeling and Reasoning in Event Calculus\\
  using Goal-Directed Constraint Answer Set Programming %
  \footnote{This paper is an extended version
    of the work by~\citet{arias19:event-calculus-asp-lopstr19}.}\thanks{%
    Work partially supported by EIT Digital, 
    MICINN projects RTI2018-095390-B-C33 InEDGEMobility
    (MCIU/AEI/FEDER, UE), PID2019-108528RB-C21 ProCode, Comunidad de
    Madrid project S2018/TCS-4339 BLOQUES-CM co-funded by EIE Funds of
    the European Union, US NSF Grants IIS 1718945, IIS 1910131, IIP
    1916206.}%
}

\author[Joaqu\'{\i}n Arias et al.]{%
Joaqu\'{\i}n Arias$^{1}$ \and Manuel Carro$^{2,3}$ \and Zhuo Chen$^{4}$ \and Gopal Gupta$^{4}$ \\
$^1$CETINIA, Universidad Rey Juan Carlos, $^2$IMDEA Software Institute\\
$^3$Universidad Polit\'ecnica de Madrid, $^4$University of Texas at Dallas\\
\texttt{\emph{joaquin.arias@urjc.es, manuel.carro@\{imdea.org,upm.es\}, \{zhuo.chen,gupta\}@utdallas.edu}}%
}

\jdate{March 2003}
\pubyear{2003}
\pagerange{\pageref{firstpage}--\pageref{lastpage}}
\doi{S1471068401001193}

\begin{document}

\makeatletter

\maketitle

\hyphenpenalty 5000 

\begin{abstract}
  Automated commonsense reasoning is essential for building human-like
  AI systems featuring, for example, explainable AI.  Event Calculus
  (EC) is a family of formalisms that model commonsense reasoning with
  a sound, logical basis.
  Previous attempts to mechanize reasoning using EC faced difficulties in
  the treatment of the continuous change in dense domains (e.g., time and
  other physical quantities), constraints among variables, default
  negation, and the uniform application of different inference
  methods, among others.
  We propose the use of s(CASP), a query-driven, top-down execution
  model for Predicate Answer Set Programming with Constraints, to
  model and reason using EC.  We show how EC scenarios can be
  naturally and directly encoded in s(CASP) and how it
  enables deductive and abductive reasoning tasks
  in domains featuring constraints involving both dense
  time and dense fluents.

  Under consideration in Theory and Practice of Logic Programming
  (TPLP)
  \vspace*{-.75em}
  
\end{abstract}

\section{Introduction}

The ability to model continuous characteristics of the world is
essential for Commonsense Reasoning (\textbf{CR}) in many domains that
require dealing with continuous change: time, the height of
a falling object, the gas level of a car, the water level in a sink,
etc.                Event Calculus (\textbf{EC}) is a formalism based on many-sorted predicate logic~\cite{kowalski,mueller_book} that can
represent continuous change and capture the commonsense law of
inertia, whose modeling is a pervasive problem in CR.  In EC,
time-dependent properties and events are seen as objects and reasoning
is performed on the truth values of properties and the occurrences of
events at a point in time.

Answer Set Programming (\textbf{ASP}) is a logic programming paradigm
that was initially proposed by~\citet{Marek99:stable_models}
and~\citet{Lifschitz1999:action_and_ASP}
to realize non-monotonic reasoning. ASP has been used by~\citet{reformulating,lee19:f2lp} to model the Event
Calculus.
Classical implementations of ASP are limited to variables ranging over
discrete and bound domains and use grounding and SAT solving
to find out models (called \textit{answer sets}) of
ASP programs.  However, reasoning on models of the real world often
needs variables ranging over dense domains (domains that are
continuous, such as $\mathbb{R}$, or that are not continuous but have
an infinite number of elements in any bound, non-singleton interval, such as
$\mathbb{Q}$).  Dense domains are necessary to accurately represent
the properties of some physical quantities, such as time, weight,
space, etc.

This paper presents an approach to modeling Event Calculus
using the s(CASP) system by~\citet{scasp-iclp2018} as the underlying
reasoning infrastructure.  The s(CASP) system is an implementation of Constraint
Answer Set Programming over first-order predicates which
combines ASP and constraints.  It features predicates, constraints among
non-ground variables, uninterpreted functions, and, most importantly, a top-down, query-driven execution strategy.  These features make it possible to return
answers with non-ground variables, possibly including constraints among
them, and to compute partial models by returning only the fragment of
a stable model that is necessary to support the answer to a given query.  Thanks to
its interface with constraint solvers, sound non-monotonic reasoning
with constraints is possible. 
This approach achieves more conciseness and expressiveness, in the
sense of being able to succinctly express complex computations and
reasoning tasks, than other related approaches.  
Dense domains can be faithfully modeled in s(CASP) as continuous
quantities, while in other proposals such
domains had to be discretized, as done by~\citet{mellarkod} and~\citet{lee19:f2lp}, therefore
losing precision or even soundness.
Additionally, in our approach the amalgamation of ASP and constraints
and its realization in s(CASP) is considerably more natural: under
s(CASP), answer set programs are executed in a goal-directed manner so
constraints encountered along the way are collected and solved
dynamically as execution proceeds --- this is very similar to the way in
which Prolog was extended with constraints. The implementation of
other ASP systems featuring constraints is considerably more complex.

In the rest of the paper we present s(CASP) and its unique
capabilities together with a terse introduction to Event Calculus
(Section~\ref{sec:background}), our approach to modeling Event
Calculus with s(CASP) (Section~\ref{sec:from-event-calculus}), a
quantitative and qualitative evaluation
(Section~\ref{sec:examples-evaluation}), and, finally, related work
and conclusions (Sections~\ref{sec:related-work}
and~\ref{sec:conclusions}).

\section{Background}
\label{sec:background}

Answer Set Programming
is a logic programming and modelling language that evaluates normal
logic programs under the stable model semantics proposed by~\citet{gelfond88:stable_models}.
s(ASP), introduced by~\citet{marple2017computing}, is a top-down,
goal-driven ASP system
that can evaluate  ASP programs with function symbols (\emph{functors})
\textbf{without} \emph{grounding} them either before or during execution.
Grounding is a procedure that substitutes  program variables with
the possible values from their domain. For most classical ASP solvers,
grounding is a necessary pre-processing phase.
Grounding, however, requires program variables to be restricted
to take values in a finite domain.  As a result, traditional ASP solvers cannot be
used to model continuous time or change.

\subsection{s(CASP)}
\label{sec:scasp}

s(CASP), presented by~\citet{scasp-iclp2018}, extends s(ASP) by adding
constraints, similarly to how CLP extends Prolog.  Also, similarly to
how s(ASP) can compute the stable model
semantics with non-ground programs and
return non-ground models including disequalities in the Herbrand
domain, s(CASP) keeps constraints as relations among variables both
during execution and in the answer sets.

Constraints have historically proved to be effective in improving both
expressiveness (programs are shorter and easier to understand, as many
computation details are taken care of by the underlying constraint
solver) and efficiency in logic programming, as they can
succinctly express properties of a solution and reduce the search
space.
As a result, s(CASP) is more expressive and faster than s(ASP), while
retaining the capability of executing non-ground predicate answer set
programs.

\subsubsection{Syntax and Behavior}
\label{sec:syntax-scasp}

An s(CASP) program is a set of clauses of the following form: \smallskip

{\centering \code{a:-c$_a$, b$_1$, $\dots$, b$_m$, not b$_{m+1}$, $\dots$, not b$_n$.} \par}

\smallskip

\noindent where \code{a} and \code{b$_1$, $\dots$, b$_n$} are
atoms.  
An atom is either a propositional variable or the expression
\code{p(t$_1$, $\ldots$, t$_n$)} if \code{p} is an $n$-ary predicate
symbol and \code{t$_1$, $\ldots$, t$_n$} are terms.
A term is either a variable \code{x$_i$} or a function symbol \code{f}
of arity \code{n}, denoted as \code{f/n}, applied to $n$ terms, e.g.,
\code{f(t$_1$, t$_2$, $\ldots$, t$_n$)}, where each \code{t$_i$} is in
turn a
term.  A function symbol of arity 0 is called a constant.  Program
variables are usually written starting with an uppercase letter, while
function and predicate symbols start with a lowercase letter.
Numerical constants are written solely with digits.\footnote{There are
  additional syntactical conventions to separate variables and
  non-variables that are of no interest here.}  Therefore, s(CASP)
accepts terms with the same conventions as Prolog: \code{f(a, b)} is a
term, and so are \code{f(g(X),Y)} and \code{[f(a)|Rest]} (to denote a
list with head \code{f(a)} and tail \code{Rest}).

{\em
\paragraph{\textbf{A note on terminology}}
ASP literature often uses the term \emph{constraint} to
denote constructions such as \code{:- p, q}, i.e., rules without
head.  They express that the conjunction of atoms %
\code{p $\land$ q} cannot be true: either \code{p},
\code{q}, or both, have to be false in any stable model.
Our  programs use as well a designated class of predicates that restrict the domains of
variables~\cite{intro_constraints_stuckey,marriott2006constraint} and
whose semantics comes from the structure and domain of an underlying
constraint system.  In Constraint Logic Programming (CLP), these
predicates are called ``constraints'' as well.  To avoid the ambiguity
that may arise from using the same name for constraints appearing
among (free) variables during program execution and in the final
models and for rules without heads, and
following~\citet{mellarkod-long}, we will refer to headless rules as
\emph{denials}.    
}

\code{c$_a$} is a simple constraint or a conjunction of constraints:
an expression establishing relations among variables in some
constraint system, as described by~\citet{intro_constraints_stuckey}.
Similar to CLP, s(CASP) is parametrized by the constraint system, from
which it inherits its semantics.
Since the execution of an s(CASP) program needs negating constraints
(Section~\ref{sec:dual}), we require that this can be done in the
constraint system by means of a finite disjunction of basic
constraints~\cite{Stuckey91,dover2000:constructive-negation}.

At least one of \code{a}, \code{b$_i$}, \code{not b$_i$}, or
\code{c$_a$} must be present.  When the head \texttt{a} is not
present it is supposed to be substituted by the head \emph{false}.
The rules have then the form

\begin{center}
\code{:-c$_a$, b$_1$, $\dots$, b$_m$, not b$_{m+1}$, $\dots$, not b$_n$.}   
\end{center}

\noindent
(and, as mentioned before, we call this rule a \emph{denial})
and their interpretation is that the conjunction of the constraints
and goals has to be false, so at least one constraint or goal has to
be false.  

The execution of an s(CASP) program starts with a \emph{query} of the
form %

\begin{center}
\code|?- c$_a$, b$_1$, $\dots$, b$_m$, not b$_{m+1}$, $\dots$, not b$_n$.|  
\end{center}

The s(SCASP) answers to a query are \emph{partial} stable models where
each one is a subset of a stable model
that satisfies the constraints,
makes non-negated atoms true, makes the negated atoms non-provable,
and, in addition, includes only atoms that are relevant to support the
query. Additionally, for each partial stable model s(CASP) can return
on backtracking the justification tree and the bindings for the free
variables of the query that correspond to the most general unifier
(\emph{mgu}) of a successful top-down derivation consistent with this
stable model.

An atom can have the form \code{-r} (i.e., have a hyphen as
its first character).  In that case it is assumed to express the
\emph{classical negation} of atom \code{r}.  Rules with head \code{-r},
to express when \code{r} is false, can 
be part of the program.  To ensure soundness, a denial 

\smallskip
\centerline{\code{:- r, -r.} }
\smallskip

\noindent
is automatically added to guarantee that atom \code{r} and its classical
negation \code{-r} are not both simultaneously true in any model.
Other than that, \code{-r} is not treated specially by s(CASP).  The
construct \code{not -r} is allowed and rules with \code{-r} in their
head or body are subject to \emph{dualization} (Section~\ref{sec:dual}).

{

  Default negation \code{not r} differs from classical negation
  \code{-r} in that \code{not r} succeeds when it cannot be proven
  from the program that \code{r} holds, while \code{-r} succeeds if
  there is a rule that states how to deduce \code{-r} and this rule,
  together with the rest of the program, can be used to derive
  \code{-r}.
  The difference from the point of view of reasoning can be
  illustrated with a simple piece of commonsense knowledge: a bus may
  cross the railway tracks if no train is approaching.  A possible
  rule using classical negation expresses would be:

\smallskip
{\centering \code{cross:- -train.} \par}
\smallskip

\noindent It means the railway tracks can be crossed if we \emph{explicitly}
know (because there is a proof for it) that no train is approaching
--- for example, because there are sensors that send us information
that ensures that there is definitely no train on the tracks
within some safety range.
The rule using default negation would be:

\smallskip
{\centering \code{cross:- not train.} \par}
\smallskip

\noindent which means that we can cross the railway tracks if there is
no evidence (because we cannot prove it) that a train is approaching
--- for example, we do not receive information of a train coming.  But
there may be no train coming, or the sensors may be faulty and not
sending signals.  That is why \code{-r} is sometimes referred to
as \emph{strong} negation: it carries with it the meaning that there
is a constructive proof that \code{r} is false.
Therefore, lack of evidence is not a hard proof.  Classical and
default negation have two different meanings inside the language and
are used to express very different common-sense reasoning scenarios.
}

When \code{-r} and \code{r} are defined, the decision to invoke \code{-r} or
\code{not r} in the body a rule depends on what the programmer wants
to express.
There is a relation of containment between \code{not r} and \code{-r},
but it is clearer in the context of non-propositional atoms.
Therefore, we defer its explanation to the end of
Section~\ref{sec:transl-narr}, when we deal with the translations of
the axioms of the Basic Event Calculus (BEC).

In s(CASP), and unlike Prolog's negation as failure and ASP default
negation, %
\code|not p(X)| can return bindings for \code{X} on success, i.e.,
bindings
for which the call \code{p(X)} would have failed.  Constraints may be
returned as well: for the program

\begin{lstlisting}[style=MyASP]
p(a).
\end{lstlisting}

\noindent
the query \code{?- not p(X).} would return the binding %
\code{X $\not=$ a} and the model \code|{not p(X $|$ {X $\not=$ a})}|,
representing the set of $not \; p(X)$ such that the atom $p(X)$ can be
proven only when $X \not= a$.\footnote{Uniqueness of
  names is assumed for constants and function names: any two constants
  or functions with different names represent
  different objects.}
Note that \code{X} in the query appeared only in a negated atom, and
did not need to be part of any non-negated atom.  This is possible
thanks to the use of constructive negation~\cite{marple2017computing}
and coinductive success~\cite{gupta2007coinductive} in s(ASP). 

These are augmented in s(CASP) with the constraint processing capabilities
presented in~\citet{scasp-iclp2018}, such that the program

\begin{lstlisting}[style=MyASP]
p(X):- X > 0.
\end{lstlisting}

\noindent
will return, for the same query as before, the model %
\code|{not p(X $|$ {X $\leq$ 0})}|.

s(CASP)
uses a top-down, goal-driven execution procedure that implements an
extension of the stable model semantics introduced
by~\citet{gelfond88:stable_models} for non-ground programs.
Default negation is solved against the \emph{dual rules} of the
program, which give a constructive definition of the negation of
program predicates.  The top-down algorithm does not need grounded
programs and makes it possible as well to return partial stable
models.

\subsubsection{Overview of the Execution Procedure of s(CASP)}
\label{sec:overview}

In the following sections we will present an abridged description of
the top-down evaluation procedure used by s(CASP), which,  in a
nutshell, is:

\begin{enumerate}
\item\label{item:consneg} Rules expressing the constructive negation
  of the predicates in the original ASP program are synthesized
  (Section~\ref{sec:dual}).  We call this the \emph{dual program}.
  Its mission is to provide a means to constructively determine the
  conditions and constraints under which calls to non-propositional
  predicates featuring variables would have failed: if we want to know
  when a rule such as \code{p(X, Y):- q(X), not r(Y).} succeeds,
      the dual program computes the constraints on \code{Y} under which
  the call \code{r(Y)} would fail.  This is an extension of the usual
  ASP semantics that is compatible with the case of programs that
  can be finitely grounded.\footnote{Note that, in the presence of
    function symbols and constraints on dense domains, this is in
    general not the case for s(CASP) programs.}
\item\label{item:negloops} The original program is checked  for loops
  of the form \code{p:- q, not p.} and denials are
  generated for them.
\item The denials generated in
  point~\ref{item:negloops}, together with any denials
  present in the original program, are collected in a
  predicate synthesized by the compiler that is invoked by adding an
  auxiliary goal to this predicate at the end of the query.
\item\label{item:interp} The union of the original program, the dual
  program, and the denials is handled by a top-down
  execution algorithm that implements the stable model semantics.
\end{enumerate}

Item number~\ref{item:interp} is specially relevant.  The dual program
(item~\ref{item:consneg}) is synthesized by means of program
transformations drawing from classical logic.  However, its meaning
differs from that of first-order logic.  That is so because it is to
be executed by a metainterpreter that does not implement the inference
mechanisms of first-order logic, as it is designed to ensure that the
semantics of answer set programs is respected.  In particular, it
treats specifically cyclic dependencies involving negation --- see
Section~\ref{sec:execution}.
For conciseness, we have not included in this paper the description of
the execution algorithm, which can be found
in~\citet{marple2012:goal-directed-asp} and~\citet{marple2017computing}.

Therefore, the soundness of s(CASP) (and its version without
constraints, s(ASP)) needs to be assessed taking into account the 
dual program, the generation of the denials and the
evaluation algorithm as a whole.  This was done
by~\citet{marple2012:goal-directed-asp} for the propositional case
(which lays the bases of the whole procedure) and extended for the
case of predicate logic, including arbitrary function symbols,
by~\citet{marple2017computing}.  We will provide reasons
supporting the soundness of the s(CASP) algorithm in the next sections.

\subsubsection{Dual Programs}
\label{sec:dual}

We summarize here the synthesis of the dual of a logic program $P$:
the completion procedure described by~\citet{Clark78} is performed to
generate a program $Comp(P)$, its rules are
converted into an equivalent form with negated heads, 
and then De Morgan's laws are applied to generate separate clauses.

\begin{enumerate}
\item Following~\citet[Section 2.1]{ferraris:stab-model-circ}, first-order
  sentences are constructed for each clause by considering each i-th
  rule of predicate $p$ 

  \begin{center}
    \code{p :- c$_a$, b$_1$, $\dots$, b$_m$, not b$_{m+1}$, $\dots$, not b$_n$.} 
  \end{center}

\noindent
  as a shorthand for %
  $\forall\vec{x}\vec{y_i} \; ( p_i(\vec{x}) \leftarrow B_i )$, where
  $B_i$ corresponds to the conjunction
  $c_a \land b_1 \land \dots \land b_m \land \neg b_{m+1} \land
  \dots \land \neg b_n$ and $\vec{y_i}$ are the variables appearing in
  $B_i$ that do not appear in $\vec{x}$.  The rationale for this
  transformation is that the grounding of an ASP program
  substitutes variables in the program clauses for all the constants in the program,
  and all the resulting clauses have to be satisfied.  This is
  precisely what the universal quantifier expresses.

  We will assume that clauses are normalized, i.e.\ head
  unifications have been made explicit as goals in the bodies of the
  corresponding clauses and the heads contain only variable names.
  Also, in what follows we will not distinguish user predicates
  \code{b$_i$} from constraints \code{c$_a$} until the last step of
  the generation of the dual program; we will make the necessary
  distinction there.

\item\label{item:and-clauses} All sentences corresponding to the
  same predicate name are conjoined together:
  \begin{align*} 
    & \forall\vec{x}\vec{y_i} \; ( p(\vec{x}) \leftarrow B_1 ) \; \land \\
    & \vdots  \\
    & \forall\vec{x}\vec{y_k} \; ( p(\vec{x}) \leftarrow B_k )
  \end{align*}

\item\label{item:or-bodies} The bodies in the antecedent of the
  sentences are joined in a single body:
  \begin{displaymath}
    \forall\vec{x}\vec{y} \; ( p(\vec{x}) \leftarrow B_1 \lor \dots
    \lor B_k) 
  \end{displaymath}
  The variables $\vec{y}$ are updated to include all $\vec{y_i}$ that
  appear in the different $B_i$ and do not appear in $\vec{x}$.
  
\item \label{item:min-quant} The scope of the quantifiers is
  minimized to make further simplifications possible:
  \begin{displaymath}
    \forall\vec{x} \; ( p(\vec{x}) \leftarrow \exists \vec{y} (B_1 \lor \dots
    \lor B_k))
  \end{displaymath}
  and then
  \begin{displaymath}
    \forall\vec{x} \; ( p(\vec{x}) \leftarrow \exists \vec{y_1} B_1 \lor \dots
    \lor \exists \vec{y_k} B_k)
  \end{displaymath}

  Transformations~\ref{item:and-clauses} to~\ref{item:min-quant} are
  valid in intuitionistic logic, and so they preserve the stable
  models of the original formulae, as mentioned by~\citet[Section
  6.1]{ferraris:stab-model-circ}.

\item\label{item:imp-to-eq} Implications are replaced by
  equivalences, to generate the Clark completion of the original
  predicate:
  \begin{displaymath}
    \forall\vec{x} \; ( p(\vec{x}) \longleftrightarrow \exists
    \vec{y_1} B_1 \lor \dots \lor \exists \vec{y_k} B_k)
  \end{displaymath}
  This transformation is, in general, not model-preserving, except in
  the case of tight programs as presented
  by~\citet{esra09:tight_programs}.  For a 
  program $P$, positive loops makes the Clark completion $Comp(P)$
  under the classical first-order semantics be weaker than $P$ under
  the stable model semantics: all stable models of $P$ are classical
  models of $Comp(P)$, but not the other way around.  Therefore, there
  may be classical models of $Comp(P)$ that are not stable models of
  $P$.

\item \label{item:create-preds} We create new predicate names to
  separate the bodies corresponding to the different original clauses:
  \begin{align*}
    &\forall\vec{x}\ (\ p(\vec{x})
        \longleftrightarrow p_1(\vec{x}) \lor \dots \lor p_k(\vec{x})\ )\\
    &\forall\vec{x}\ (\ p_i(\vec{x}) \longleftrightarrow \exists \vec{y_i}  B_i)
  \end{align*}

  \item Their  duals $\lnot p/n$, $\lnot p_i/n$ are:
  \begin{align*}
    & \forall\vec{x}\ (\ \lnot p(\vec{x}) \longleftrightarrow 
        \lnot ( p_1(\vec{x}) \lor \dots \lor p_k(\vec{x}))\ )\\
    & \forall\vec{x}\ (\ \lnot p_i(\vec{x}) \longleftrightarrow
        \lnot \ \exists \vec{y}_iB_i\ )
  \end{align*}
  This is a semantically-preserving operation in the classical logic
  semantics, and so the models of the Clark completion remain
  untouched.

\item De Morgan's laws are applied to the first formula (the ``entry
  point'' of the negated predicate) and the existential quantifier is
  negated:
    \begin{align*}
& \forall\vec{x}\ (\ \lnot p(\vec{x}) \longleftrightarrow 
    \lnot p_1(\vec{x}) \land \dots \land \lnot p_k(\vec{x})\ )\\
& \forall\vec{x}\ (\ \lnot p_i(\vec{x}) \longleftrightarrow
  \forall \vec{y}_i \neg B_i\ )
  \end{align*}
  De Morgan's Law are of course semantics-preserving in classical
  logic, but also in the stable model semantics because the formulas
  $\neg(A \land B)$ and $\neg A \lor \neg B$ are strongly
  equivalent, as mentioned by~\citet[Proposition 5]{lifschitz01:strong_equivalence}.

\item For each dual rule corresponding to each $p_i$, an auxiliary
  \emph{negated} predicate corresponding to the \emph{negated} body is
  synthesized:
  \begin{align*}
    & \forall\vec{x}\ (\ \lnot p(\vec{x}) \longleftrightarrow
     \lnot p_1(\vec{x}) \land \dots \land \lnot p_k(\vec{x})\ )\\
    &\forall\vec{x}\ (\ \lnot p_i(\vec{x}) \longleftrightarrow
     \forall \vec{y}_i \neg p^\prime_i(\vec{x},\vec{y}_i)\ ) \\
    &\forall\vec{x}\vec{y}\ (\ \lnot p^\prime_i(\vec{x},\vec{y})
     \longleftrightarrow \neg B_i\ )
  \end{align*}
  Let us remember that each $B_i$ has the form
  $b_{i,1} \land \dots \land b_{i,m} \land \lnot\ b_{i,m+1} \land
  \dots \land \lnot\ b_{i,n}$. Therefore, the last sentence is 
  \begin{displaymath}
    \forall\vec{x},\vec{y}\ (\ \lnot p^\prime_i(\vec{x},\vec{y})
     \longleftrightarrow \neg (b_{i,1} \land \dots \land b_{i,m} \land
     \lnot\ b_{i,m+1} \land  \dots \land \lnot\ b_{i,n})\ ) 
  \end{displaymath}

\item We apply De Morgan's Law to the last sentence in the previous
  point to obtain
    \begin{align*}
    &\forall\vec{x},\vec{y}\ (\ \lnot p^\prime_i(\vec{x},\vec{y})
     \longleftrightarrow \neg b_{i,1} \lor \dots \lor \neg b_{i,m} \lor
     b_{i,m+1} \lor  \dots \lor b_{i,n}\ ) 
    \end{align*}
    
  \item We revert the introduction of the equivalence.
    This transformation changes the models of program w.r.t.\ that of
    the Clark completion.  However, programs under the stable
    semantics (and under Prolog semantics) have a clear notion of
    direction
            and the metainterpreter only uses the goal-driven direction ``use
    \emph{Body} to prove \emph{Head}''.  The closed-world assumption,
    captured by Clark's completion, is already implicit in the top-dow
    evaluation algorithm of s(CASP).
        \begin{align*}
    & \forall\vec{x}\ (\ \lnot p(\vec{x}) \leftarrow
     \lnot p_1(\vec{x}) \land \dots \land \lnot p_k(\vec{x})\ )\\
    &\forall\vec{x}\ (\ \lnot p_i(\vec{x}) \leftarrow
     \forall \vec{y}_i \neg p^\prime_i(\vec{x},\vec{y}_i)\ ) \\
    &\forall\vec{x}\vec{y}\ (\ \lnot p^\prime_i(\vec{x},\vec{y})
     \leftarrow \neg b_{i,1} \lor \dots \lor \neg b_{i,m} \lor
     b_{i,m+1} \lor  \dots \lor b_{i,n})\ ) 
    \end{align*}

  \item Separate the disjunction in the body of the last sentence in
    different clauses:
 \begin{align*}
    &\forall\vec{x}\vec{y}\ (\ \lnot p^\prime_i(\vec{x},\vec{y})
      \leftarrow \neg b_{i,1}\ ) \\
    &\forall\vec{x}\vec{y}\ (\ \lnot p^\prime_i(\vec{x},\vec{y})
      \leftarrow \neg b_{i,2}\ ) \\
    & \vdots \\
    &\forall\vec{x}\vec{y}\ (\ \lnot p^\prime_i(\vec{x},\vec{y})
      \leftarrow b_{i,n}\ ) 
    \end{align*}

    These clauses, together with the original program and the
    denials, are used by the metainterpreter to
    decide whether some atom belongs or not to a stable model of a
    program and to return the (minimal) support for that atom.

\end{enumerate}
  
\medskip \noindent This provides a definition for $ \lnot p(\vec{x})$
via a clause with head $\lnot p_i(\vec{x})$ for each original clause
with head $p_i(\vec{x})$.
The newly introduced negated atoms $\neg b_{i.1} \ldots \neg b_{1.m}$
can fall into two categories: they are either negations of user
predicates or negations of constraints.  In the former case, the
procedure just described generates a definition for the negation of
user predicates.  In the latter case, constraints are either head
unifications created after normalizing the clauses, or they are
constraints in a different domain.  Both cases are treated similarly:

\begin{itemize}
\item If the negation of a constraint can be expressed as a finite
  disjunction of basic constraints~\cite{dover2000:constructive-negation,Stuckey91}, the
  compiler makes that expansion.  In the simplest case, that
  disjunction is a single constraint: a linear constraint $E_1 < E_2$
  is translated into $E_1 \geq E_2$.  In other cases, it can be a
  ``real'' disjunction: the constraint $E_1 = E_2$ in
  CLP($\mathbb{Q}$) (linear constraints over the rationals), is
  negated by converting it into $E_1 < E_2 \lor E_1 > E_2$, and each
  component of the disjunction is handled by a clause.
        In practice, it is not necessary that all constraints in a constraint
  system can be negated, but only those that are required in a given
  program to answer some query.

\item When the negation of a constraint cannot be expressed as a
  finite disjunction of constraints, we make a best effort to provide
  an \emph{ad-hoc} implementation.  For example, for the equalities
  $v_i = t_i$ in the Herbrand domain CLP($\mathcal{H}$) that were
  added when normalizing clauses, we introduce a call
  $\mathit{diff}(v_i, t_i)$ to a disequality solver provided by
  the runtime environment.  Negation of equality in CLP($\mathcal{H}$)
  can be expressed as a finite disjunction only for programs that can
  generate a finite number of ground terms.
\end{itemize}

Executable code for the dual program is generated by removing the
external quantifiers (as in Horn clauses) and translating the
universal quantifiers that were applied to local variables into
a call to the predicate \code{forall(Var, Pred)}, provided by the
s(CASP) runtime.

  \begin{example}
  \label{ex:dual}
    Given the program

    \vspace{-1em}

    \begin{multicols}{2}
\begin{lstlisting}[style=MyASP]
p(X):- q(X, Z), not r(X).  
p(Z):- not q(X, Z), r(X).
q(X, a):- X .>. 5.  
r(X):- X .<. 1.
\end{lstlisting}
    \end{multicols}

    \vspace{-1em}
    
    \noindent
    its dual is shown below
    
    \begin{multicols}{2}
\begin{lstlisting}[style=MyASP, basewidth=.48em]
% not p/1
not p(A) :- not p_1(A), not p_2(A).

not p_1(A) :- forall(B, not p_1(A,B)).
not p_1(A,B) :- not q(A,B).
not p_1(A,B) :- r(A).

not p_2(A) :- forall(B, not p_2(A,B)).
not p_2(A,B) :- q(B,A).
not p_2(A,B) :- not r(B).
% not q/1
not q(A,B) :- not q_1(A,B).

not q_1(A,B) :- B \= a.
not q_1(A,B) :- A #=< 5.

% not r/1
not r(A) :- not r_1(A).
not r_1(A) :- A #>= 1.
\end{lstlisting}
    \end{multicols}
    
        \end{example}

\subsubsection{Sketch of the Execution Scheme}
\label{sec:execution}

Queries to the original program extended with the dual rules are
evaluated by a runtime environment.  This is currently a
metainterpreter written in Prolog that executes the algorithm
described by~\citet{marple2012:goal-directed-asp}.  This algorithm has
similarities with SLD resolution, but it takes into account specific
characteristics of ASP and the dual programs, such as the different
kinds of loops, the denials, and the introduction of universal
quantifiers in the body of the clauses.
The main highlights of this algorithm are:

\begin{description}
\item[Loop handling:]

  Two different cases are distinguished by~\citet{marple2012:goal-directed-asp}:
 \begin{itemize}
  \item When a call eventually invokes itself and there is an odd number
    of intervening negations (as in, e.g., %
    \quad \mbox{\code{p:- q. $~$ q:- not r. $~$ r:- p.}}), the evaluation
    fails (and backtracks) to avoid contradictions of the form
    $p \land \lnot p$.
  \item When there is an even number of intervening negations, as in 
        \mbox{\code{p:- not q. $~$ q:- r. $~$ r:- not p.}}
       the metainterpreter generates several stable models, such as
    \code|{p, not q, not r}| and \code|{q, r, not p}|.
  \end{itemize}

\item[Denials:]
  The s(CASP) compiler automatically generates an auxiliary
  predicate that captures all the denials
  written by the programmer.
  This predicate is invoked during query evaluation to ensure that the
  returned models are consistent with the denials.  The
  current implementation executes them at the end of the query
  evaluation, when a candidate model has been generated.  It would
  however be possible to check them at appropriate points while the
  execution proceeds, in order to increase
  performance, as suggested by~\citet{marple14:dynamic}.

  The s(CASP) compiler also detects statically rules of the form %
  \code{r:-q, not r.} and introduces denials to ensure
  that the models satisfy $\lnot q \lor r$, even if the atoms
  \code{r} or \code{q} are not needed to solve the query.
  This is done by building a dependency graph of the program and
  detecting the paths where this may happen, including across several
  calls.  For the propositional case, such an analysis can be precise.
  For the non-propositional case, 
    an over-approximation is calculated.  In both cases, denials that
  are not used during program evaluation can be generated.  These
  may impose a penalty in execution time, but are safe.  Therefore,
  s(CASP) will state that the program

  \vspace{-1em}

  \begin{multicols}{3}
\begin{lstlisting}[style=MySCASP]
p :- not q.
q :- not p.
r :- not r.
\end{lstlisting}
  \end{multicols}
  
  \vspace{-2em}
  
  \noindent
  has no stable models, regardless of the initial query. 

\item[Universal quantification:] 
  Universal quantifications in the body of the clauses are translated
  into the construction \code{forall(Var, Pred)}.  This is implemented
  by the runtime environment by solving \code{Pred}, extracting the
  constraints attached to the quantified variables, and using these
  constraints negated to narrow the constraint store under which
  \code{Pred} is executed.  This is iterated until failure or until
  the constraint store has an empty domain for the quantified
  variables.  \citet{scasp-iclp2018} present this algorithm in more
  detail.

\end{description}

\subsubsection{Execution with Unsafe Variables and Uninterpreted
  Function Symbols}
\label{sec:unsafe}

The code in Example~\ref{ex:dual} has variables that would be termed
as \emph{unsafe} in regular ASP systems: variables that appear in
negated atoms in the body of a clause, but that do not appear in any
positive literal in the same body.  Since s(CASP) synthesizes explicit
constructive goals for these negated goals, the aforementioned code
can be run as-is in s(CASP).  The query %
\code{?- p(A).} generates three different models:

\begin{lstlisting}[style=MySCASP, numbers=none]
{ p(A | {A $>$ 5}), q(A | {A $>$ 5}, a), not r(A | {A $>$ 5}) }
  A $>$ 5

{ p(A | {A $\not=$ a}), not q(B | {B $<$ 1}, A | {A $\not=$ a}), r(B | {B $<$ 1}) } 
  A $\not=$ a

{ p(a), not q(B | {B $<$ 1}, a), r(B | {B $<$ 1}) } 
  A = a
\end{lstlisting}

\noindent
where the notation \code'V | {C}' for a variable \code{V} is intended
to mean that \code{V} is subject to the constraints in \code|{C}|.
The constraints \code{A = 5}, \code{A $\not=$ a}, and \code{A = a}
correspond to the bindings of variable \code{A} that make the atom
in the query \code{?- p(A)}  belong to the stable model.

Another very relevant point where s(CASP) differs from ASP is in the
possibility of using arbitrary uninterpreted function symbols to
build, for example, data structures.  While in mainstream ASP
implementations these could give rise to an infinite grounded program,
the s(CASP) execution model can deal with them similarly to Prolog, with
the added power of the use of constructive negation in the execution
and in the returned models.

\begin{example}
  \label{ex:using-lists}
  The predicate \code{member/2} below models the membership to a list
  as it is usual in (classical) logic programming.  The query is
  intended to derive the conditions for one argument not to belong to
  a given list.

\begin{lstlisting}[style=MySCASP]
member(X, [X|Xs]).  
member(X, [_|Xs]):- member(X, Xs).

list([1,2,3,4,5]).

?- list(A), not member(B, A).
\end{lstlisting}
\noindent
This program and query return in s(CASP) the following model and binding:

\begin{lstlisting}[style=MySCASP, numbers=none]
{ list([1,2,3,4,5]),
   not member(B | {B $\not=$ 1,B $\not=$ 2,B $\not=$ 3,B $\not=$ 4,B $\not=$ 5}, [1,2,3,4,5]), 
   not member(B | {B $\not=$ 1,B $\not=$ 2,B $\not=$ 3,B $\not=$ 4,B $\not=$ 5}, [2,3,4,5]),
   not member(B | {B $\not=$ 1,B $\not=$ 2,B $\not=$ 3,B $\not=$ 4,B $\not=$ 5}, [3,4,5]),
   not member(B | {B $\not=$ 1,B $\not=$ 2,B $\not=$ 3,B $\not=$ 4,B $\not=$ 5}, [4,5]), 
   not member(B | {B $\not=$ 1,B $\not=$ 2,B $\not=$ 3,B $\not=$ 4,B $\not=$ 5}, [5]),
   not member(B | {B $\not=$ 1,B $\not=$ 2,B $\not=$ 3,B $\not=$ 4,B $\not=$ 5}, []) }
   A = [1,2,3,4,5], B $\not=$ 1, B $\not=$ 2, B $\not=$ 3, B $\not=$ 4, B $\not=$ 5
\end{lstlisting}
\end{example}

\noindent
I.e., for variable \code{B} not to be a member of the list
\code{[1,2,3,4,5]} it has to be different from each of its elements.

In addition to default negation, s(CASP) supports classical negation
to capture the explicit evidence that a literal is false, as mentioned
in Section~\ref{sec:syntax-scasp}.

s(CASP) is implemented in Ciao
Prolog~\cite{hermenegildo11:ciao-design-tplp-medium} and is available
at \mbox{\url{https://gitlab.software.imdea.org/ciao-lang/scasp}}.

\subsubsection{s(CASP) as a Conservative Extension of ASP}
\label{sec:semantics}

The behavior of s(CASP) and ASP is the same for propositional
programs.
For programs featuring unsafe variables (legal in ASP, but not in
mainstream ASP systems) or programs that could create data structures
arbitrarily large or whose variable ranges are defined in infinite 
domains (either unbound or bound but dense), which are outside the
standard domain of ASP systems as they cannot be finitely
grounded, 
s(CASP) extends ASP in a consistent way.  The domain of the
variables is implicitly expanded to include a domain which can be
potentially infinite.

Let us use an example introduced
in~\cite[Pag.12]{marple2017computing}.
%
We are interested in knowing whether \code{p(X)} (for some
\code{X}) is or not part of a stable model:

\begin{lstlisting}[style=myASP]
d(1).
p(X) :- not d(X).
\end{lstlisting}

The only constant in the program is \code{1}, which is the only
possible domain for \code{X} in the second clause.  That clause is not
legal for ASP, as \code{X} is an unsafe variable (Sect.~\ref{sec:unsafe}).
Adding a domain
predicate call for it (i.e., adding \code{d(X)} to the body of the
second clause), makes its model be \code|{d(1)}| (\code{not p(1)}
is implicit).

That second clause is however legal in s(CASP).  Making the query
\code{?-p(X)} returns the \emph{partial} model %
\code'{p(X|{X \= 1}), not d(X|{X \= 1})}' stating that \code{p(X)} and
\code{not d(X)} are true when %
\code{X \= 1}, which is consistent with, but more general than, the
model given by ASP.  As the model is partial, only the atoms (perhaps negated)
involved in the proof for \code{?- p(X)} appear in that model.

\subsection{Circumscription}
\label{sec:circ}

\emph{Circumscription}~\cite{mccarthy80:circumscription,Lifschitz85:computing-circ}
is a technique to perform non-monotonic reasoning within the framework
of first-order logic.  Circumscription minimizes the
extension\footnote{The \emph{extension} of a predicate is the set of
  tuples for which predicate is true.}  of the predicates that we want
to circumscribe.  Intuitively, it aims at formalizing that the known
objects in a certain class are \emph{all} the objects that are in that
class.  Event Calculus theories require that some of their
predicates
are circumscribed to ensure that they can only be interpreted as they
appear in the description of the scenario.

The following definition of circumscription is
due to~\citet{Lifschitz85:computing-circ}:

\begin{definition}[Circumscription]
  Let $A(P, Z)$ denote a sentence, where $P$ is a tuple of predicate
constants and $Z$ a tuple of function and/or predicate constants
disjoint with $P$.
The circumscription of $P$ in $A(P, Z)$ with
variable $Z$ is defined as the second order sentence
$A(P, Z) \land \neg\exists p, z (A(p, z) \land p \mathbf{<} P)$.
The $\mathbf{<}$ symbol in the previous expression is defined as follows:
if $U$ and $V$ are n-ary predicates, $U \mathbf{\leq} V$ stands for
$\forall x_1 \ldots x_n (U(x_1,\ldots , x_n ) \rightarrow V(x_1,
\ldots, x_n))$. $U = V$ stands for $U \mathbf{\leq} V$ and
$V \mathbf{\leq} U$ and $U \mathbf{<} V$ stands for
$U \mathbf{\leq} V \land \neg(V \mathbf{\leq} U )$.  These definitions
are extended to tuples of predicates in the obvious fashion.
\end{definition}

$U \mathbf{\leq} V$ expresses that the extension of $U$ is a subset of
the extension of $V$, and $U \mathbf{<} V$ means that the extension of
$U$ is a proper subset of the extension of $V$.  The circumscription
of $A(P, Z)$ is a formula whose extension is the minimal extension of
the predicates in $P$ that makes $A(P, Z)$ true, and in which the
objects in $Z$ are allowed to vary.  It is expressed with
$CIRC(A; P; Z)$ or $CIRC(A; P)$ if $Z$ is empty.

\begin{example}
  Let's take $A = P(a)$. Its minimization should express that $P$ is
  only true for $a$:
  \begin{displaymath}
    CIRC(A; P) \equiv \forall x (P(x) \leftrightarrow x = a)
  \end{displaymath}
\end{example}

\begin{example}
  Let's take $A = \neg P(a)$. We only have information about $P$ not
  being true in $a$, which is the only constant that appears in $A$,
  but also \emph{all} the constants that appear in the sentence. Its
  minimization  expresses this as:
  \begin{displaymath}
    CIRC(A; P) \equiv \forall x \neg P(x)
  \end{displaymath}
\end{example}

Computing circumscriptions is in general a hard task.  However, there
are class of formulas (\emph{separable formulas}) for
which it has shown to be easy by~\citet{Lifschitz85:computing-circ}.
Also,~\citet{ferraris:stab-model-circ} have shown that there is a close
relationship between circumscription and stable models and
indeed~\citet[Definition 2]{reformulating} proved that the stable
model semantics coincides with the circumscription for a large class
of formulas called \emph{canonical formulas}.  Non-canonical formulas
can often be rewritten as canonical formulas, therefore expanding the
range of coincidence of circumscription and stable model semantics.

\subsection{Event Calculus}
\label{sec:event-calculus}

\begin{figure}[t]
  \centering
  \begin{tabular}{p{0.27\linewidth}p{0.52\linewidth}}

      \textbf{Predicate} & \textbf{Meaning}\\
      \midrule
$\mathit{InitiallyN}(f)$       & fluent $f$ is false at time 0   \\ 
$\mathit{InitiallyP}(f)$       & fluent $f$ is true at time 0    \\ 
$\mathit{Happens}(e, t)$       & event $e$ occurs at time $t$      \\
$\mathit{Initiates}(e, f, t)$  & if $e$ happens at time $t$, $f$ is true and
                     not released from the commonsense law of inertia
                     after $t$  \\ 
$\mathit{Terminates}(e, f, t)$ & if $e$ occurs at time $t$, $f$ is false and not
                      released from the commonsense law of inertia
                      after $t$ \\ 
$\mathit{Releases}(e, f, t)$   & if $e$ occurs at time $t$, $f$ is released from the
                      commonsense law of inertia after $t$ \\ 
$\mathit{Trajectory}(f_1, t_1, f_2, t_2)$ & if $f_1$ is initiated by an event that
                             occurs at $t_1$, then $f_2$ is true at $t_2$ \\ 
      \noalign{\vspace {.25cm}}
$\mathit{StoppedIn}(t_1, f, t_2)$ & $f$ is stopped between $t_1$ and $t_2$ \\
      $StartedIn(t_1, f, t_2)$ & $f$ is started between $t_1$ and $t_2$ \\
      \noalign{\vspace {.25cm}}      
$\mathit{HoldsAt}(f, t)$       & fluent $f$ is true at time $t$    \\

    \end{tabular}
    \caption{Basic event calculus (BEC) predicates}
\centerline{($e$ = event, $f$, $f_1$, $f_2$ = fluents, $t$, $t_1$, $t_2$ = timepoints)}
    \label{fig:predicates}
\end{figure}

EC (presented at length by, for example,~\citet{mueller_book}) is a
formalism for reasoning about events and change, of which there are
several axiomatizations.  There are three basic, mutually related,
concepts in EC: \emph{events}, \emph{fluents}, and \emph{time points}.
An event is an action or incident that may occur in the world: for
instance, a person dropping a glass is an event.
A fluent is a time-varying property of the world, such as the altitude
of a glass.  
A time point is an instant in time.  
Events may happen at a time point; fluents have a truth value at any
time point or over an interval, and their truth values are subject to
change upon the occurrence of an event.  In addition, fluents may have
(continuous) quantities associated with them when they are true.

For example, the status of a glass falling may be represented by two
fluents: one that captures the fact that the glass is falling and
another one that captures the height of the glass over the ground.
The \emph{event} of dropping a glass initiates the \emph{fluent} that
captures that the glass is falling and gives some initial value to its
height.  This height changes with time according to some formula.  The
event of catching the glass makes the fluent that reflects it is
falling false and makes its height not to change, possibly until a further
event takes place.
An EC description consists of a domain narrative
(Fig.~\ref{fig:predicates}) and a universal theory
(Fig. \ref{fig:BEC-axioms}).  The domain narrative consists of the
causal laws of the domain, the known events, and the fluent
properties, and the universal theory is a conjunction of EC axioms
that encode, for example, the inertia laws.

The original EC (OEC) was introduced by~\citet{kowalski}.  OEC has sorts for event occurrences,
fluents, and time periods.
In this paper we use the Basic Event Calculus (BEC) formulated by~\citet{shanahan-ec-explained-99}
 as presented by~\citet{BEC}. BEC allows fluents to be released from the
commonsense law of inertia via the $Release$ predicate, and adds the
ability to represent continuous change via the $Trajectory$ predicate.
Fig.~\ref{fig:BEC-axioms} summarizes the seven axioms of the BEC
theory.  An explanation of these axioms follows:

\begin{figure}[t]
  
\newcommand{\Axiom}[3]{\textbf{#1} & #2 &\\
   \multicolumn{3}{r}{#3} \\[0.75em]}
\newcommand{\AxiomB}[3]{\textbf{#1} & \multicolumn{2}{r}{#2 \hfill #3}\\[0.75em]}

\centering
\begin{tabular}{llr}

    \Axiom{BEC1.}%
    {$\mathit{StoppedIn}(t_1,f,t_2) \ \ \equiv $}%
    {$\exists e,t\ \  (\ \mathit{Happens}(e,t) \land t_1<t<t_2 \land (\ \mathit{Terminates}(e,f,t) \lor \mathit{Releases}(e,f,t)\ )\ )$}
    \Axiom{BEC2.}%
    {$\mathit{StartedIn}(t_1, f, t_2) \ \ \equiv $}%
    {$\exists e, t \ \ (\ \mathit{Happens}(e, t) \land t_1 < t < t_2 \land (\ \mathit{Initiates}(e, f, t) \lor \mathit{Releases}(e, f, t)\ )\ )$}
    \Axiom{BEC3.}%
    {$\mathit{HoldsAt}(f_2, t_2) \ \ \leftarrow$}%
    {$\ \ \ \ $  $\mathit{Happens}(e, t_1) \land \mathit{Initiates}(e, f_1, t_1) \land \mathit{Trajectory}(f_1, t_1, f_2, t_2) \land \lnot \mathit{StoppedIn}(t_1, f_1, t_2)$} 
    \AxiomB{BEC4.}%
    {$\mathit{HoldsAt}(f, t) \ \ \leftarrow$}%
    {$ \mathit{InitiallyP}(f ) \land \lnot \mathit{StoppedIn}(0, f, t)$}
    \AxiomB{BEC5.}%
    {$\lnot \mathit{HoldsAt}(f, t) \ \ \leftarrow$}%
    {$\mathit{InitiallyN}(f ) \land \lnot \mathit{StartedIn}(0, f, t)$}
    \Axiom{BEC6.}%
    {$\mathit{HoldsAt}(f, t_2) \ \ \leftarrow$}%
    {$ \mathit{Happens}(e, t_1) \land \mathit{Initiates}(e, f, t_1) \land t_1 < t_2 \land \lnot \mathit{StoppedIn}(t_1, f, t_2)$}
    \Axiom{BEC7.}%
    {$\lnot \mathit{HoldsAt}(f, t_2) \ \ \leftarrow\ \ $}%
  {$ \mathit{Happens}(e, t_1) \land \mathit{Terminates}(e, f, t_1) \land t_1 < t_2 \land \lnot \mathit{StartedIn}(t_1, f, t_2)$}

  \end{tabular}
  \caption{Formalization of BEC theory as presented by~\protect\citet{mueller_book}.}
\label{fig:BEC-axioms}
\end{figure}

\begin{itemize}
\item \textbf{Axiom BEC1}. A fluent $f$ is stopped between time points $t_1$ and
$t_2$ iff it is terminated or released by some event $e$ that occurs after
$t_1$ and before $t_2$.

\item \textbf{Axiom BEC2}. A fluent $f$ is started between time points $t_1$ and
$t_2$ iff it is initiated or released by some event $e$ that occurs after
$t_1$ and before $t_2$.

\item \textbf{Axiom BEC3}. A fluent $f_2$ is true at time  $t_2$ if
a fluent $f_1$ initiated at $t_1$ does not finish before $t_2$ and it
makes fluent $f_2$ be true.\footnote{For implementation convenience,
  and without loss of expressiveness, we assume that argument $t_2$ in
  $Trajectory(f_1, t_1, f_2, t_2)$ is not a time difference w.r.t.\
  $t_1$, but an absolute time after $t_1$.}

\item \textbf{Axiom BEC4}. A fluent $f$ is true at time $t$ if it is true at
time 0 and is not stopped on or before $t$.

\item \textbf{Axiom BEC5}. A fluent $f$ is false at time $t$ if it is false at
time 0 and it is not started on or before $t$.

\item \textbf{Axiom BEC6}. A fluent $f$ is true at time $t_2$ if it is initiated
at some earlier time $t_1$ and it is not stopped before $t_2$.

\item \textbf{Axiom BEC7}. A fluent $f$ is false at time $t_2$ if it is
terminated
at some earlier time $t_1$ and it is not started on or before
$t_2$.

\end{itemize}

\section{From Event Calculus to s(CASP)}
\label{sec:from-event-calculus}

\subsection{Circumscription in s(CASP)}

Circumscription
is applied to EC domain narratives, and as a result, the events that
happen and their effects are only those explicitly defined.
As mentioned before, the definition of a given scenario (its
\emph{narrative} part) states the basic actions and effects using the
predicates in Fig.~\ref{fig:predicates}.
Let us consider example 14 by~\citet{mueller_book}, which reasons about
the turning on and off of a light switch:

\begin{displaymath}
  \begin{array}{ll}
    \mathit{Happens}(e,t)\ \  \equiv & (e=\mathit{TurnOn} \land t=2)\ \ \lor \\
                        &  (e=\mathit{TurnOff} \land t=4)
  \end{array}
\end{displaymath}

Assuming circumscription, we can write the axioms $\mathit{Happens}(\mathit{TurnOn},2)$
and $\mathit{Happens}(\mathit{TurnOff},4)$, instead of writing the previous
formula, while ensuring that there are no events (resp., effects)
other than those stated.  I.e., we can prove that the light is off at
$t=6$, because we can prove the absence of an event turning the light
on between $t=4$ and $ t=6$.  If we need to add a new event (e.g.,
$\mathit{Happens}(\mathit{TurnOn},5)$) we only have to add it, instead of modifying the
formula that expresses the circumscription of the narrative.  It has
to be noted that in this case the circumscribed formula is simple, but
the circumscription of more complex formulas is more involved.

Similarly, since EC assumes the circumscription of the rest of the
predicates defined in the narrative ($\mathit{InitiallyN}$,
$\mathit{InitiallyP}$, $\mathit{Initiates}$, $\mathit{Terminates}$,
$\mathit{Releases}$, and $\mathit{Trajectory}$), we have that the
explicitly known effects of events are the only effects of events.

EC does not assume circumscription for the underlying theory. Since
BEC1 and BEC2 are definitions which can be expanded wherever they
appear, they do not need to be taken care of specially.

However, for the rest of the BEC theory axioms in
Fig.~\ref{fig:BEC-axioms}, from BEC3 to BEC7, we \textbf{cannot}
apply this assumption because they are not circumscribed and they are
implications: $\mathit{HoldsAt}$ is true \emph{if} the body of one of its
corresponding axioms holds. This means that, if we cannot infer that
$\mathit{HoldsAt}(f,t)$ is true, then, we \textbf{cannot} deduce that
$\lnot \mathit{HoldsAt}(f,t)$ is true.
Instead, the falsehood of $\mathit{HoldsAt}(t,f)$ must be inferred through the
axioms BEC5 and BEC7 if this is supported by direct evidences defined
in the narrative.
It is important to note that if the narrative describes a scenario in
which it is possible to deduce that a fluent $f$ is true and false at
the same point in time $t$, this narrative is inconsistent and there
are no valid models for this scenario. Furthermore, if a given
narrative is described in terms of the truth or falsehood of a fluent
$f$ at some point in time $t$, but we are not able to decide this
value, then there are multiple valid models because we have that
$\mathit{HoldsAt}(t,f) \lor \lnot \mathit{HoldsAt}(t,f)$.

Let us describe below how we use s(CASP) to compute the
circumscription based on predicate
completion~\cite{mueller_book,reformulating} while we are able to
express the truth/falsehood of $\mathit{HoldsAt}$ using classical negation.

\subsection{Modeling BEC with s(CASP)}
\label{sec:scasp-as-target}

Two key factors contribute to s(CASP)'s ability to model Event
Calculus: the preservation of non-ground variables 
during the execution and the integration with constraint solvers.

\medskip
\noindent\textbf{Treatment of variables in s(CASP):}
Thanks to the use of non-ground variables, s(CASP) is able to
directly model Event Calculus axioms that would otherwise require
``unsafe'' rules (Section~\ref{sec:unsafe}).
Let us take, for example, rule BEC4 of Fig.~\ref{fig:BEC-axioms}.  In
a straightforward encoding (see Fig.~\ref{fig:BEC_in_sCASP}), the
parameter $t$ (\code{T} in the code), which appears in the head and
does not appear in a positive literal in the body (i.e., it only
appears in $\lnot \mathit{StoppedIn}(0, f, t)$) would be classified as unsafe
by a mainstream ASP system.

It may be argued that a way to overcome this issue would be to
translate the negation in the axioms as classical negation, i.e.,
\code{-stoppedIn(0, F, T)}.  However, this would need to generate a
negation of the two clauses of \code{stoppedIn/3} which would in turn 
need the (classical) negations of \code{terminates/3}, \code{happens/2},
and \code{releases/3}.  However, these narrative predicates do not
come with a definition stating when they are false, in the classical
sense of negation.

An ASP solver such as \emph{clingo}~\cite{clingo} will not be
able to directly process unsafe rules like this. The standard approach
to fix unsafe rules is to add a positive literal defining the domain
of the unsafe variable (\code{T} in this case), but this is not
feasible if we want to maintain the property that \code{T} represents
time and is therefore not discrete and not finite.
On the other hand, the top-down execution strategy of s(CASP) makes it
possible to keep logical variables both during execution and in answer
sets and therefore free (logical) variables can be handled in heads
and in negated atoms.
	
\medskip\noindent\textbf{Integration with constraint solvers:}
The s(CASP) system has a generic interface to enable plugging in
constraint solvers.  s(CASP) currently includes the CLP($\mathbb{Q}$) linear constraints solver by~\citet{holzbaur-clpqr},
that supports the arithmetic constraints $<, >, =, \leq, \geq$.  This
is used to implement the definitions and axioms of BEC that require
comparisons of points in (continuous) time and to solve the equations
that arise from these comparisons.
The selection of CLP($\mathbb{Q}$) instead of the faster
CLP($\mathbb{R}$) is motivated by soundness reasons.  Since
CLP($\mathbb{R}$) uses internally floating-point numbers, rounding and
approximations compromise accuracy and termination of some code.  On
the other hand, CLP($\mathbb{Q}$) represents rational numbers exactly
and therefore it should not introduce any calculation error.  One
example in which the use of floating-point numbers would be inadequate
is the code in Fig.~\ref{fig:tap}, which uses the factor
$\frac{4}{3}$ and that does not have an exact floating-point
representation.

The s(CASP) infrastructure is however parametric w.r.t.\ the
underlying constraint solver, and other implementations of constraint
domains can be implemented and plugged in if necessary.  As an
example, s(CASP) includes a solver for disequality in the Herbrand
domain (Sec.~\ref{sec:dual}), which is necessary to generate the dual
of almost any interesting program.

\begin{figure}[t]
\begin{multicols}{2}
\begin{lstlisting}[style=myASP, belowskip=-2pt, basewidth=.52em]
%% BEC1
stoppedIn(T1,F,T2) :- 
    T1.<.T, T.<.T2, 
    terminates(E,F,T),
    happens(E,T).  

stoppedIn(T1,F,T2) :-
    T1.<.T, T.<.T2,
    releases(E,F,T),
    happens(E,T).

%% BEC2
startedIn(T1,F,T2) :-
    T1.<.T, T.<.T2,
    initiates(E,F,T),
    happens(E,T).

startedIn(T1,F,T2) :-
    T1.<.T, T.<.T2,
    releases(E,F,T),
    happens(E,T).

%% BEC3
holdsAt(F2,T2) :-
    initiates(E,F1,T1), 
    happens(E,T1),
    trajectory(F1,T1,F2,T2), 
    not stoppedIn(T1,F1,T2).
%% BEC4
holdsAt(F,T) :-
    0.<.T, 
    initiallyP(F),
    not stoppedIn(0,F,T).

%% BEC5
-holdsAt(F,T) :- 
    0.<.T, 
    initiallyN(F), 
    not startedIn(0,F,T).

%% BEC6
holdsAt(F,T2) :- 
    T1.<.T2, 
    initiates(E,F,T1), 
    happens(E,T1), 
    not stoppedIn(T1,F,T2).

%% BEC7
-holdsAt(F,T2) :- 
    T1.<.T2, 
    terminates(E,F,T1), 
    happens(E,T1),
    not startedIn(T1,F,T2).

%% Consistency (automatically added)
:- -holdsAt(F,T), holdsAt(F,T).
\end{lstlisting}
  \end{multicols}
  
  \caption{Basic Event Calculus (BEC) modeled in s(CASP)}
  \label{fig:BEC_in_sCASP}
  
\end{figure}

\subsection{Translating the BEC Axioms into s(CASP)} 

Our translation of the BEC axioms into s(CASP) is related to the
translation used by the systems EC2ASP and
F2LP~\cite{reformulating,lee19:f2lp}.
We differ in three aspects that improve performance for a top-down
system, fully use s(CASP)'s ability to treat unbound variables, and
do not compromise the soundness of the translation (according to the
proofs presented by~\citet{reformulating}). These are:
\textit{the treatment of rules with negated heads, the possibility of
  generating unsafe rules,} and \textit{the use of constraints over
  dense domains} (rationals, in our case).
We describe below, with the help of a running example, the translation
that turns logic statements (as found in BEC) into an s(CASP) program.
The code corresponding to the translations of the axioms of BEC in
Fig.~\ref{fig:BEC-axioms} can be found in Fig.~\ref{fig:BEC_in_sCASP}.
s(CASP) code follows the syntactical conventions of logic programming:
constants (including function names) and predicate symbols start with
a lowercase letter and variables start with an uppercase letter. In
addition, numerical constraints are written as constraints in s(CASP),
(e.g., \code{.<.}) to make it clear that they do not correspond to
Prolog's arithmetic comparisons:

\begin{itemize}
\item \textbf{Atoms and Constants:} Their names are preserved.  \emph{Uniqueness of
  Names}~\cite{shanahan-ec-explained-99} is assumed by default (and
enforced) in logic programming.

\item \textbf{Constraints:} Predicates that represent constraints
  (e.g., on time) are directly translated to their counterparts in
  s(CASP).  E.g., $t_1 < t_2$ becomes \mbox{\code{T1 .<. T2}}, which is handled
  by the CLP($\mathbb{Q}$) solver.
      The translation
  is parameterized on the constraint
  domain. 

\item \textbf{Definitions:}
  The axiomatization of BEC uses definitions of the form
  $D(x) \equiv \exists y B(x,y)$, where $B(x,y)$ is a conjunction of
  (possibly negated) atoms, disjunctions of atoms, and constraints
  (e.g., BEC1).  The use of these definitions
    makes it easier to build and reuse conceptual blocks out of basic
  predicates.   
  They are however not strictly necessary \emph{per se} as predicates.
    For performance reasons
      we treat them as if they were written as
  $\forall x (D(x) \leftarrow \exists y B(x,y))$,
  following the work of~\citet{lee19:f2lp}.  Since $D$, as mentioned above, is
  given a name as a convenience to write the BEC axioms, we can ignore
  its truth value in the (partial) models that s(CASP) generates
  because if it were expanded where it is used, it would have
  disappeared.  Therefore, the models returned using
  implication or equivalence for the definition of $D$ are the same.


\item \textbf{Rules with Positive Heads:} A rule (e.g., BEC6)
\begin{displaymath}
  \forall x  (H(x) \leftarrow \exists y (A(y) \land  \neg B(x, y) \land x < y))
\end{displaymath}
where $x < y$ is a constraint, is translated into
\begin{lstlisting}[style=myASP]
h(X) :- X.<.Y, a(Y), not b(X,Y).
\end{lstlisting}
The constraint \code{X.<.Y} could be placed anywhere in the clause.
However, in top-down evaluation schemes, the general recommendation is
to execute constraints as soon as possible to use a \emph{constraint
  and generate} mechanism instead of \emph{generate and test} in order
to improve performance.  In the very common case where user-level
constraint operations are translated into constraint
propagation, which is usually required to be deterministic,
executing them earlier in the tree simplifies the constraints without
performing internal search or creating search branches.  On the
other hand, by constraining the domains of the variables earlier, the
size of the search trees needed by calls to user predicates (\code{a(Y), not b(X, Y)} in this case) are reduced.

\item \textbf{Rules with Negated Heads:} BEC rules 5 and 7 infer
  negated heads $\neg \mathit{HoldsAt}(f, t)$ while rules 4 and 6 infer positive
  heads $\mathit{HoldsAt}(f, t)$, i.e., they follow, respectively, the scheme
\begin{displaymath}
\forall x  (\neg H(x) \leftarrow \exists y A(x, y)) \ \ \land \ \ \forall x
 (H(x) \leftarrow \exists y B(x, y))
\end{displaymath}
The standard approach to translate rules with negated heads is to
convert them into denials~\cite{reformulating} such
as \code{:- a(X,Y), h(X)}.
Our approach is to create instead the atom \code{-h(X)} to denote
(Section~\ref{sec:syntax-scasp}) the negation of \code{h(X)} and a rule
that captures the explicit evidence that \code{h(X)} is false:
\begin{lstlisting}[style=my ASP]
-h(X) :- a(X,Y).
\end{lstlisting}
This makes it possible to invoke \code{-h(X)} as a regular predicate
in a top-down execution.
The compiler will additionally (Section~\ref{sec:syntax-scasp}) generate
denials
to ensure that \code{-h(X)} and \code{h(X)} cannot be simultaneously
true.  Therefore, s(CASP) will detect an inconsistency
if both $\mathit{HoldsAt}(f, t)$ and $\neg \mathit{HoldsAt}(f, t)$ can be
simultaneously derived from the narrative.
Having rules stating explicitly when %
$\neg \mathit{HoldsAt}(f, t)$ can be derived makes it possible to query for it,
as some BEC rules need.  We will later see how this is connected
with the translation of the narrative.

\item \textbf{Rules with Disjunctive Bodies:} A rule (e.g., BEC1)
\begin{displaymath}
    \forall x  [H(x) \leftarrow \exists y (\ (A(x,y) \lor B(x,y)) \land C(x,y)\ )]
\end{displaymath}
is translated into two separate clauses:
\begin{lstlisting}[style=myASP]
h(X) :- a(X,Y), c(X,Y).
h(X) :- b(X,Y), c(X,Y).
\end{lstlisting}

\end{itemize}

\subsection{Translation of  the Narrative}
\label{sec:transl-narr}

Every basic BEC predicate $P(x)$ from the narrative
(Fig.~\ref{fig:predicates}) 
is translated into s(CASP) as a set of rules of the form 
\begin{displaymath}
 P(x) \leftarrow \gamma
\end{displaymath}
where the body $\gamma$ is a conjunction of atoms, negated atoms, and constraints
that can be trivially $true$ in some cases.  The rules corresponding
to the same $P(x)$ are assumed to state \textbf{all} the cases where
$P(x)$ is true. %

Throughout this section we will use the light scenario problem
from~\citet[example 14]{mueller_book}: a two-color bulb lamp can be
switched on and off.  We want to be able to answer when the light is
on (and its color in this case) and off.  We will present the BEC
narrative and its translation into s(CASP), available in full in
Figure~\ref{fig:light}.

\begin{figure}[t]
\begin{multicols}{2}
\begin{lstlisting}[style=myASP, belowskip=-2pt, basewidth=.45em]
happens(turn_on, 2).
happens(turn_off, 4).
happens(turn_on, 5).

initiates(turn_on, on, T).

terminates(turn_off, on, T).
terminates(turn_off, red, T).
terminates(turn_off, green, T).
trajectory(on, T1, red, T2) :- 
    T1.<.T2, 
    T2.<.T1+1.
trajectory(on, T1, green, T2) :- 
    T2.>=.T1+1.

releases(turn_on, red, T).  
releases(turn_on, green, T).
\end{lstlisting}
  \end{multicols}
  
  \caption{Narrative of the light scenario modeled in s(CASP)}
  \label{fig:light}
  
\end{figure}

\paragraph{\textbf{Events:}}
Let us consider the description below:  \begin{displaymath}
  \begin{array}{ll}
    \mathit{Happens}(e,t)\ \  \equiv & (e=\mathit{TurnOn} \land ( t=2 \lor t=5) )\ \ \lor \\
                        &  (e=\mathit{TurnOff} \land t=4)
  \end{array}
\end{displaymath}

It states that the $\mathit{TurnOn}$ event will happen at time $t=2$ and
$t=5$, and that $\mathit{TurnOff}$ will happen at $t=4$.  As we
mentioned before, since EC assumes circumscription, it is equivalent
to the axioms $\mathit{Happens}(\mathit{TurnOn},2)$, $\mathit{Happens}(\mathit{TurnOff},4)$ and
$\mathit{Happens}(\mathit{TurnOn},5)$, which are translated as facts (lines 1-3 of
Fig.~\ref{fig:light}).

\paragraph{\textbf{Event Effects:}}
The effects of events are represented using the predicates
$Initiates(e,f,t)$  and $Terminates(e,f,t)$.
In our example, when the event $\mathit{TurnOn}$ happens, the light is put in
$on$ status; similarly, when the event $\mathit{TurnOff}$ happens, the light
will not be $on$, $red$ or $green$:
\begin{displaymath}
  \begin{array}{lll}
  \mathit{Initiates}(e,f,t) \ & \equiv & (e=\mathit{TurnOn} \land f = \mathit{On})\\
  \mathit{Terminates}(e,f,t)\ & \equiv & (e=\mathit{TurnOff} \land (f=\mathit{On} \lor f=\mathit{Red} \lor f=\mathit{Green}))
  \end{array}
\end{displaymath}

In both cases, this can happen at any time $t$, and the translation
again becomes facts (lines 5-9 in Fig.~\ref{fig:light})

\paragraph{\textbf{Release from Inertia:}}
When turned on, the light emits red light for one second and
after that it starts to emit green light.  $Trajectory$ expresses how
this change depends on the time elapsed since an event occurrence.

\begin{displaymath}
  \begin{array}{lll}
  \mathit{Trajectory}(s,t_1,c,t_2)\ & \equiv & (s = \mathit{On} \land c = \mathit{Red} \land
                                      t_1<t_2 \land t_2<t_1 + 1) \; \lor \\
  & & (s = \mathit{On} \land c = \mathit{Green}
                                              \land t_2 \geq t_1 + 1)
  \end{array}
\end{displaymath}

The translation is in lines 10-14 of Fig.~\ref{fig:light}.
$\mathit{Releases}$ states that the color of the light is released from the
commonsense law of inertia. After a fluent is released, its truth
value is not determined by BEC and it can change. Thus, there
may be models in which the fluent is true, and models in which the
fluent is false: \begin{displaymath}
  \begin{array}{lll}
  \mathit{Releases}(e,f,t)\ & \equiv & (e=\mathit{TurnOn} \land (f=\mathit{Red} \lor f=\mathit{Green}))
  \end{array}
\end{displaymath}
Note that the time parameter $t$ appears only in the head (it is
universally quantified).
Releasing a fluent (lines 16 and 17 in Fig.~\ref{fig:light}) frees it
up so that other axioms in the domain description can be used to
determine its truth value, thus allowing us to represent continuous
change of the fluent.

\paragraph{\textbf{State Constraints:}}
State constraints usually contain $\mathit{HoldsAt}(f,t)$ or
$\lnot \mathit{HoldsAt}(f,t)$ and represent restrictions on the models.  In our
running example, a light cannot be red and green at the same time:
\begin{displaymath}
  \begin{array}{l}
    \lnot \mathit{HoldsAt}(\mathit{Red},t) \lor \lnot \mathit{HoldsAt}(\mathit{Green},t)
  \end{array}
\end{displaymath}
Note that this is logically equivalent to
$\forall t \ \lnot (\mathit{HoldsAt}(\mathit{Red},t) \land \mathit{HoldsAt}(Green,t))$ and is
translated as \mbox{\code|:-holdsAt(red,T),holdsAt(green,T)|}.
Adding this denial to the program in Fig.~\ref{fig:light} does not
change its models. However, if we change the trajectory definition for
the red light stating $t_2 \leq t_1 + 1$ instead of $t_2 < t_1 + 1$
the state constraint is violated at $t_2 = t_1 + 1$ and therefore,
there are no valid models.

\paragraph{\textbf{A Note on Using $\neg \mathit{HoldsAt}(f,t)$ in the Narrative:}}
The narrative predicates may depend on what the BEC theory can
deduce.  In other words, $\mathit{HoldsAt}(f,t)$ or $\neg \mathit{HoldsAt}(f,t)$ can
(perhaps indirectly) be used in the body $\gamma$ of a narrative rule
(see an example in Fig.~\ref{fig:tap}).  $\mathit{HoldsAt}(f,t)$ can be invoked
directly, but $\neg \mathit{HoldsAt}(f,t)$ ought to be called using classical
negation, e.g., \code{-holdsAt(F,T)}.  As we mentioned before, the
reason is that circumscription is not applied to the EC axioms and we
can deduce only the truth (or falsehood) of $\mathit{HoldsAt}$ when we have
direct evidence of either of them --- i.e., what the positive
(\code{holdsAt(F,T)}) and negative (\code{-holdsAt(F,T)}) heads
provide.

As we mentioned before, the consistency rule (line 56 in
Fig.~\ref{fig:BEC_in_sCASP}) introduced by the compiler of s(CASP)
would ensure that \code{holdsAt(F,T)} and \code{-holdsAt(F,T)} are
mutually exclusive by flagging an inconsistency.
Note that if \code{not holdsAt(F,T)} succeeds,
then \code{holdsAt(F,T)} is false but it does not imply
that \code{-holdsAt(F,T)} is true (i.e.,
\code{-holdsAt(F,T) $\Rightarrow$ not holdsAt(F,T)}, but it is
\textbf{not} the case that %
\code{not holdsAt(F,T) $\Rightarrow$ -holdsAt(F,T)}). 
Symmetrically for \code{not -holdsAt(F,T)} we have that %
\code{holdsAt(F,T) $\Rightarrow$ not -holdsAt(F,T)}.

In previous implementations of EC, such as F2LP, reasoning about the
falsehood of $\mathit{HoldsAt}(f,t)$ can be made using only the default
negation, implemented as negation as failure (i.e.,
\code{not/1}). Therefore, the presence of classical negation in
s(CASP) not only increases the expressive power from the point of view
of the programmer (by, e.g., determining whether a fluent is or not
active at some point in time), but it also ensures correctness in
those cases where %
\code{-holdsAt(F,T) $\not\equiv$ not holdsAt(F,T)}, as inconsistent
models are discarded.

\subsection{Continuous Change: A Complete Encoding}
\label{sec:cont-chang-compl}

\begin{figure}[t]
  \begin{multicols}{2}
\begin{lstlisting}[style=myASP, basewidth=.48em]
#include bec_theory.

max_level(10) :- not max_level(16).
max_level(16) :- not max_level(10).

initiallyP(level(0)).

happens(overflow,T).
happens(tapOn,5).

initiates(tapOn,filling,T).  
terminates(tapOff,filling,T).
initiates(overflow,spilling,T):- 
    max_level(Max),
    holdsAt(level(Max), T).  

releases(tapOn,level(0),T):-
    happens(tapOn,T).
trajectory(filling,T1,level(X2),T2):- 
    T1.<.T2,
    X2.=.X+4/3*$\ap$T2-T1$\cp$, 
    max_level(Max),
    X2.=<.Max,
    holdsAt(level(X),T1).  
trajectory(filling,T1,overlimit,T2):-
    T1.<.T2,
    X2.=.X+4/3*$\ap$T2-T1$\cp$, 
    max_level(Max),
    X2.>.Max,
    holdsAt(level(X),T1).  

trajectory(spilling,T1,leak(X),T2):-
    holdsAt(filling, T2), 
    T1.<.T2, 
    X.=.4/3*$\ap$T2-T1$\cp$.
\end{lstlisting}
  \end{multicols}

  \caption{Encoding of an Event Calculus narrative with continuous change}
  \label{fig:tap}

\end{figure}

We consider now an extension of the water tap example
by~\citet{shanahan-ec-explained-99}, where we define two possible
worlds and added a triggered fluent to describe the ability of s(CASP)
to model complex scenarios.
In this example, a water tap fills a vessel, whose capacity is either
10 or 16 units, and when the level of water
reaches the bucket rim, it starts spilling.  Let us present the main
ideas behind its encoding, available in Fig.~\ref{fig:tap}.

\paragraph{\textbf{Continuous Change:}}
The fluent $\mathit{Level}(x)$ represents that the water is at level
$x$ in the vessel.
The first $Trajectory$ formula (lines 19-24) determines the
time-dependent value of the $\mathit{Level}(x)$ fluent, which is active as long
as the $Filling$ fluent is true and the rim of the vessel is not
reached.
The second $Trajectory$ formula (lines 25-30) allows us
to capture the fact that the water reached the rim of the vessel and
overflowed.
Note the $\frac{3}{4}$ factor that relates time and water amount.  As
mentioned before, if the underlying solver approximates the numerical
operations, water levels could be miscalculated and therefore the
answers to some queries could be wrong.  The use
of CLP($\mathbb{Q}$) prevents this.

\paragraph{\textbf{Uniqueness of $\mathit{Level}(x)$}}

A relevant question is whether 
the fluent $\mathit{Level}(x)$ could take two different values at the same time.
Intuitively, it should not, because if we are modeling faithfully a
physical system evolving under a series of events, the level should be
unique at any point in time.  Note, however, that if this were to
happen, it would be because the narrative does not correspond to the
reality.  The specification given by~\citet{shanahan-ec-explained-99}
includes explicitly an axiom 

\begin{displaymath}
  \mathit{HoldsAt}(\mathit{Level}(x_1),t) \land \mathit{HoldsAt}(\mathit{Level}(x_2),t) \rightarrow x_1 = x_2
\end{displaymath}
to avoid this situation.

We did not include it for two reasons: on the one hand, a careful
inspection of the narrative reveals that this could not be the case.
On the other hand, if this axiom were necessary, the model without it
would allow the simultaneous existence of two alternative water
levels.  That raises the question whether the model is correct, as the
narrative would in that case allow two different, diverging events
happen at the same time, or the trajectories would allow two
inconsistent fluents overlap.  Just stating this inconsistency is of
little value in practice, as it does not help catch errors in the
model and it removes possible correct states.%
\footnote{Let us note that EC does not preclude a fluent to have
  different associated values simultaneously: it is only the semantics
  of this fluent that may disallow it. As an example, let us take 
  the fluent $\mathit{Occupied}(n)$ that expresses that the seat $n$
  in a theater is occupied.  Obviously, several instances for
  different $n$ can be true at the same time.}

\paragraph{\textbf{Triggered Fluent:}}
The fluent $Spilling$ is triggered (lines 13-15) when the water
level reaches the rim of the vessel.  As a consequence, 
the $Trajectory$ formula in lines \mbox{32-35} starts the fluent $Leak(x)$
and captures the amount of water leaked while the fluent
$Spilling$ holds.

\paragraph{\textbf{Alternative Worlds:}}
As we mentioned in Section~\ref{sec:execution}, the presence of even
loops generates different worlds. In our implementation, the clauses
in lines 3-4 force the vessel capacity to be either 10 or 16, and
therefore, they create two possible worlds/models:
\code|{max_level(10), not$\ $max_level(16),$ \dots$}| and %
\code|{max_level(16), not max_level(10),$ \dots$}|.

Different worlds can be used to model alternative scenarios where an event
may happen in one world and not in another.  For this, a keyword
\code{#abducible} is provided as a shortcut in s(CASP).  We will use
it in Sec.~\ref{sec:abduction}, below.

\section{Examples and Evaluation}
\label{sec:examples-evaluation}

The benchmarks
used in this section are available at
\url{http://www.cliplab.org/papers/EC-sCASP-TPLP2020/}. They
were run on a macOS 10.15.7 laptop with an Intel Core i5 at
2GHz.

\subsection{Deduction}
\label{sec:deduction}

Deduction determines whether a state of the world is possible given
a theory
and an initial narrative.
We can perform deduction in BEC for the previous examples through
queries to the corresponding s(CASP) program.  For the \emph{lights} scenario
(Fig.~\ref{fig:light}):

\begin{description}
\item \code{?-holdsAt(on,3)} succeeds: it states that the
  light is on at time 3.
\item \code{?--holdsAt(on,4.5)} succeeds: the light is not on at
  time 4.5.\footnote{The decimal number 4.5 is automatically converted to
    rational representation.}
\item \code{?-holdsAt(F,3)} is true in one stable model containing
  \code{holdsAt(green,3)} and \code{holdsAt(on,3)}, meaning that the
  light is on and green at time 3.
\end{description}

\vspace{.5em}

\noindent
Additionally, as we mentioned in Section~\ref{sec:transl-narr}, using
the default negation \code{not/1} we can check the absence of a proof
for \code{holdsAt/2} and \code{-holdsAt/2}. However, there are time
points, e.g., at time 1, where neither the truth nor the falsehood for
the fluent representing that the light is on can be deduced from the
program. Therefore, the queries \code{?- not holdsAt(on,1)} and
\code{?- not -holdsAt(on,1)} would both succeed.

Finally, let us use the water level scenario (Fig.~\ref{fig:tap}) to
make queries involving time and the water level (that are continuous
physical quantities):
\begin{description}
\item \code{?-holdsAt(level(H),15/2)} is true when \code{H=10/3}.
\item \code{?-holdsAt(level(10/3),T)} is true when \code{T=15/2}.
\end{description}

\vspace{.5em}

Note that, as explained with more detail in the \emph{Evaluation}
subsection below, s(CASP) can operate and answer correctly queries
involving rationals (and, in general, dense domains as long as they
are supported by the underlying constraint solver) without having to
modify the original program to introduce domains for the relevant
variables or to \emph{scale} the constants to convert rationals into
integers.

\subsection{Abduction}
\label{sec:abduction}

Abductive reasoning can be used to determine a sequence of events/actions that reaches a given state.
In the case of ASP, actions are naturally captured as the set of atoms
that are true in a model that includes the initial and final states,
and that are consistent with the BEC theory.
For the water scenario (Fig.~\ref{fig:tap}), let us assume we want to
determine whether the
water can reach a level of 12 at time 14. The query
\mbox{\code|?-holdsAt(level(12),14)|} will return a single model with
a vessel size of 16 and the rest
of the atoms in the model capturing what must (not) happen to reach
this state.  For EC, the relevant atoms are those related to the
events that happen (as these trigger fluents according to the model rules
in \code{initiates/3}, \code{releases/3}, and \code{trajectory/4})
and, to keep track of the state of the system, the atom
\code{holdsAt/2} and its classical negation.  If, for the query
mentioned above, we restrict the model to these atoms,\footnote{This
  can be done automatically by placing \code{#show} 
  directives in the source code.} we obtain the following filtered model:

\begin{lstlisting}[style=MySCASP, numbers=none]
{  initiallyP(level(0)), not happens(tapOn,D | {D #> 0,D #< 5}),
   holdsAt(level(0),5), happens(tapOn,5),
   not happens(tapOff,F | {F #> 5,F #< 14}), holdsAt(level(12),14) }
\end{lstlisting}

\noindent
where we have listed the atoms in increasing order of time stamp.

This subset of the model states what must happen to reach a state with
a water level of 12 at time $t = 14$.  It captures events that must be
observed (e.g., \code{happens(tapOn,5)}, meaning that the water tap
has to open at time $t = 5$), what is the observable state of the
fluents (\code{holdsAt(level(12),14)}, meaning that the water level at $t = 14$
is 12), but also what events \textbf{must not} happen %
(\code+not happens(tapOn, D | {D #> 0,D #< 5})+, meaning that the tap must not be
opened between $t = 0$ and $t = 5$, or %
\code+not happens(tapOff,F | {F #> 5,F #< 14})+, meaning that the
water tap must not be closed between $t = 5$ and $t = 14$).

Additionally, since s(CASP) only generates partial models, it does not
contain atoms that express actions that are not necessary for the
conclusion, i.e., the plan does not contain references to actions
(either positive or negative) that do not interfere with the final
state.  For example, it does not state that the event of closing the
tap must or must not happen between $t = 0$ and $t = 5$: whether
this event happens or not is immaterial for the final
result. 
Other abductive tasks can be performed:
adding the directive \code{#abducible} to the fact
\code{happens(tapOn,5)}, we specify that it is possible (but not
necessary) for the tap to be open at time \code{5}. As we mentioned in
Section~\ref{sec:cont-chang-compl}, this directive is translated into
code that creates different worlds/models.  For the previous query,
\code{?-holdsAt(level(L),14)} (that determines the level of water at
$t = 14$),
we obtain two alternative partial models:
\begin{itemize}

\item One containing the literal
  \code|happens(tapOn,5)|
  meaning that the tap is open at $t = 5$, and therefore, the
  resulting model is the previous one.

\item Another one containing
  \code|{ holdsAt(level(0),14), initiallyP(level(0)), not happens(tapOn,G $|$ {G #> 0,G #< 5}),  not happens(tapOn,5), not happens(tapOn,E $|$ {E #> 5,E #< 14}) } |
  meaning that the tap is \textbf{not} open at $t = 5$ (and neither
  for $0 < t < 5$, nor for $5 < t < 14$), and
  therefore, the water level at $t = 14$ remains equal to 0, which causes the 
   literal \code{holdsAt(level(0),14)} be part of the model.
\end{itemize}

Note that s(CASP) determined the truth value of $Happens$ and, more
importantly, performed constraint solving to infer the time ranges
during which some events ought (and ought not) to take place,
represented by the negated atoms in the models inferred by
constructive negation. Since all relevant atoms have a time parameter,
they actually represent a \emph{timed plan}.  Due to the
expressiveness of constraints, this plan contains information on time
points when events must (not) happen and also on time \emph{windows}
(sometimes in relation with other events) during which events must
(not) take place.  Note that it would be impossible to (finitely)
represent this interval with ground atoms, as it corresponds to an
infinite number of points.

\subsection{Evaluation}
\label{sec:evaluation}

A direct performance comparison of our implementation of BEC in
s(CASP) with implementations in other systems may not be meaningful:
most previous systems implement \emph{discrete} Event Calculus (DEC)
and they do not support continuous quantities.  Since offering this
support is one of the key points of our proposal, giving up on it and
comparing with an implementation DEC in s(CASP) is pointless and
defeats the main purpose of this piece of work.  We will then have to
compare BEC in s(CASP) with how a system that implements DEC can be
used to approximate the results we can give.

One of the ASP-based tools that support DEC is F2LP, an ASP-based system that according to~\citet{reformulating} \emph{``outperforms
\emph{DEC reasoner}~\cite{mueller2008discrete}''}, reported
there as the more efficient SAT-based implementation.
F2LP is a tool that executes DEC by turning first-order formulas under
the stable model semantics into a logic program without constraints
that is then evaluated using an ASP solver.

Our first evaluation compares the light scenario in
Fig.~\ref{fig:light} running under s(CASP) with the F2LP translation
under \emph{clingo 5.1.1}, the current version of the state-of-the-art
ASP system. Since the directive \code{#domain} is no longer available
in \emph{clingo}, we adapted the translation of F2LP adding
\code{timestep(1..10)} and \code{timestep/1} to make the clauses safe
(Appendix A of the supplementary material accompanying the paper at
the TPLP archive).  While under s(CASP) we can reason about time
points in an unbounded, dense domain, the encoding used by F2LP makes
time belong to the integers (in particular, to the interval from 1 to
10, with the previous \texttt{timestep/1} fact).  Therefore, since the
light can be red for $t>2, t<3$ and for $t>5, t<6$,\footnote{When
  turned on (at $t=2$ and $t=5$), the light emits red light for one
  second (see Section~\ref{sec:transl-narr}).}
there are no \textbf{integer} time points between 1 and 10 where the
emitted light is red.  Therefore, the query \code{?-holdsAt(red,T)}
does not return any model under \emph{clingo}, while the execution of
the same query under s(CASP) returns the constraint conjunctions
\code{T.>.2, T.<.3} and \code{T.>.5, T.<.6}.

In order to determine at what time point the red light is on using
\emph{clingo}, we modified the program generated by F2LP to
refine the \code{timestep} domain with 
\code{timestep(1..10*P):-  precision(P)}, 
where the new predicate \code{precision(P)} makes it possible to have
a finer grain for the possible values of \code{timestep} by increasing
the value of \code{P}.
With this modification, it is possible to check if the light is red at
time $t=5.9$ with \emph{clingo} by stating that we want a precision of
tenths (using \code{precision(10)}) and modifying the queries
accordingly, e.g., using \mbox{\code|?-holdsAt(red,59)|}.  Similarly,
using \code{p=100} in \code{precision(P)} it is possible to check for
$t=5.99$ by querying \mbox{\code|?-holdsAt(red,599)|}, and so on.

This change (see Appendix B of the supplementary material accompanying
the paper at the TPLP archive, for the complete program) is in
principle not difficult to perform, but it undoubtedly obfuscates the
resulting program (and for more complex narratives it would be harder,
impractical, or even infeasible), and also impacts negatively its
performance.
Table~\ref{tab:experiment-a} shows how the additional precision
necessary in the F2LP encoding increases the execution run-time of \emph{clingo} by orders of
magnitude. On the other hand, s(CASP) does not have to adapt its
encoding/queries and its performance does not change.

\begin{table}[tb]
  \caption{Comparative table of s(CASP) and F2LP+clingo.}

    \vspace{1em}

    {  \renewcommand{\code}{\lstinline[style=myASP]}

  \begin{subtable}{.90\linewidth}
  \caption{Run time (ms) comparison for the light scenario.}
  \label{tab:experiment-a}
    \centering
    \begin{tabular}{p{5cm}rp{.5cm}rc}
      \toprule
      Queries    & \multicolumn{1}{p{1cm}}{s(CASP)} & &
 \multicolumn{2}{p{4cm}}{F2LP+clingo and precision} \\
      \midrule
      \code|?-holdsAt(red,5.9)|      & 228 & & \textbf{82} & ~~10 \\
      \code|?-holdsAt(red,5.99)|      & \textbf{240} & & 8,364 & ~100\\
      \code|?-holdsAt(red,5.999)|     & \textbf{226} & & $>$ 5 min. & 1000\\
            \bottomrule        
    \end{tabular}
  \end{subtable}
  
\vspace{1em}

  \begin{subtable}{.90\linewidth}
  \caption{Run time (ms) comparison for the water tap scenario.}
  \label{tab:experiment-b}
    \centering
    \begin{tabular}{p{5cm}rp{.5cm}rc}
      \toprule
      Queries    & \multicolumn{1}{p{1cm}}{s(CASP)} & &
     \multicolumn{2}{p{4cm}}{F2LP+clingo and precision} \\
      \midrule
      \code|?-holdsAt(level(11),T)|    & \textbf{301} & & 475 & ~~10* \\    
          &  & & 77,305 & 100 \\    
      \bottomrule        
    \end{tabular}
    \centerline{(*) With this precision the value returned by \emph{clingo}
      is wrong.}
\end{subtable}
  }

\end{table}

The second benchmark is the water level scenario in Fig.~\ref{fig:tap}.
The physical model needs continuous quantities but, as it
happened in the previous case, 
its execution with \emph{clingo} forces the level of water to be
expressed using integers. 
For example, the query \mbox{\code{?-holdsAt(level(11),T)}} returns
the value \code{T=53/4} (= 13.25) using s(CASP),\footnote{s(CASP)
  returns a rational number because it uses the rational constraint
  solver CLP($\mathbb{Q}$) in order not to lose precision, as would
  happen if using floating-point numbers. s(CASP) can however output
  them in decimal notation by using the command-line flag \code|-r|.}
while the execution under \emph{clingo} fails:
the level of water is $h = 10.\overline{6}$ at time $t = 13$ and
$h = 12$ at time $t = 14$; therefore there is no integer point in time
$t$ with a level of water $h = 11$.

As before, we modified the program generated by F2LP by adding the
predicate \code{precision(P)} to specify the precision of the program
by scaling the numbers \emph{clingo} deals with, which is necessary 
to determine at what time the level of the water is $11$.
For this example, the resulting encoding (see Appendix C of the
supplementary material accompanying the paper at the TPLP archive) is
more complex than for the previous benchmark.  Additionally, this
example shows that the scaling value depends on the particular
benchmark: \code{precision(10)} is not fine-grained enough to capture
the solution, and we have to go up to \code{precision(100)} to obtain
a model with \code{T=1325} (corresponding to $t=13.25$, the correct
value) for the query \code{holdsAt(level(1100),T)}.

An undesirable effect of rounding in ASP is that rounding may not only
make programs fail, but it may make them succeed with wrong answers.
For example, in the water level example, with \code{precision(10)},
the query \code{holdsAt(level(110),T)} holds with \code{T=133} (which
would correspond to $t = 13.3$).  This value is not right, and it is
due to arithmetic rounding performed during the program execution.

Also, note that increasing \code{P} as powers of \code{10}, as we have
done, may not always work: scaling units to make an answer such as
\code{1/3} expressible as integers needs the scaling factor of
\code{3} included.  That points to the need to have some
knowledge about the program answers \emph{before} scaling the program,
or to transform the whole program by using multipliers containing
e.g. all the denominators that can appear at run-time during
the program execution.

Table~\ref{tab:experiment-b} shows that, also for this benchmark, the
additional precision increases the execution run-time of \emph{clingo}
by orders of magnitude. 

\bigskip

\section{Related Work}
\label{sec:related-work}

Previous work translated \emph{discrete} EC into
ASP by 
reformulating the EC models as first-order stable models and
translating the (almost universal) formulas of EC into a logic program
that preserves stable models.  Given a finite domain, EC2ASP (and its
evolution, F2LP) compiles (discrete) Event Calculus formulas into
ASP programs~\cite{lee19:f2lp,reformulating}.  This translation
scheme relies on two facts: the semantics of second-order circumscription and first-order stable models coincide on canonical formulas, and
almost-universal formulas can be transformed into a logic program
while preserving the stable models.  As a result, computing models
of Event Calculus descriptions can be done by computing the stable
models of an appropriately generated program.

Clearly, approaches featuring discrete domains cannot faithfully
handle continuous quantities such as time. In addition, because of
their reliance on SAT solvers to find the stable models, they can only
handle \textit{safe} programs. In contrast, in the s(CASP) system,
because of its direct support for predicates with arbitrary terms,
constructive negation, and the novel \textit{forall
  mechanism}~\cite{marple2017computing,scasp-iclp2018}, program safety
is not a requirement.  Thus, s(CASP) can model Event Calculus axioms
much more directly, elegantly, and in continuous domains.

The approaches mentioned above assume discrete quantities and do not
support reasoning about continuous time or change.  As long as
SAT-based ASP systems are used to model Event Calculus, continuous
fluents cannot be straightforwardly expressed since they require
unbound or dense domains for the variables. The work closest to incorporating continuous time makes use of SMT solvers.  In
this approach, constraints are incorporated into ASP and the grounded
theory is executed using an SMT solver, as in~\citet{aspmt}. However, this
approach
has not been directly applied to modeling the Event Calculus. The
closest tool chain is ASP Modulo Theory to SAT Modulo Theory
(ASPMT2SMT) by~\citet{DBLP:conf/jelia/BartholomewL14}
that uses \emph{gringo} to partially ground the ASPMT theories
and generate constraints that are processed by \emph{Z3}~\cite{z3}.
However, regular, discrete ASP variables are at the heart of the
model, and these are grounded and used to generate the constraints.
Therefore, if these discrete variables approximate continuous
variables in the model, the constraints generated will only
approximate the conditions of the original problem and therefore their
solutions will also be an approximation (or a subset) of the solutions
for the real problem. In other words, the initial discretization done
for the ASP variables will be propagated via the generated constraints
to the final solutions that will, in the best case, be a discretized
version of the actual solutions.  As an example, if time is
discretized, the solutions to the model will suffer from this
discretization.

EC can be written as a (Horn-clause) logic program, but it cannot be
executed directly by Prolog, as reported
by~\citet{shanahan_abductive}, because it lacks some necessary
features, such as constructive negation, deduction of negated atoms,
and (to some extent) detection of infinite failure.  The last one can
to some degree be worked around by using variants of Prolog
implementations that feature loop-breaking mechanisms in the
presence of constraints, as done by~\citet{TCLP-tplp2019}.  The other
points may eventually need ad-hoc coding for every example or crafting
an interpreter able carry out deduction tailored to the task.

A common approach is to write a metainterpreter specific to the EC
variant at hand.  This can be as complex as writing a (specialized)
theorem prover or, more often, a specialized interpreter whose
correctness is difficult to ascertain (see the code
by~\citet{chittaro96:eff-temp-reasoning-CEC}).
Therefore, some Prolog implementations of EC do not completely
formalize the calculus or implement a weaker version.  In our case,
we leverage the capabilities of s(CASP) to provide constructive,
sound negation,  negative rule heads, and loop
detection.

\vspace{-.5em}

\section{Conclusions}
\label{sec:conclusions}

We showed how Event Calculus can be modeled in s(CASP), a
goal-directed implementation of constraint answer set programming with
predicates, with  fewer limitations than other approaches. s(CASP)
can capture the notion of continuous time (and, in general, fluents)
in Event Calculus thanks to its grounding-free top-down evaluation
strategy. It can also represent complex models and answer queries in a
flexible manner thanks to the use of constraints.

The main contribution of the paper is to show how Event Calculus 
can be directly modeled using s(CASP), an ASP system that
seamlessly supports constraints. The modeling of Event Calculus using
s(CASP) is more elegant and faithful to the original axioms compared
to other approaches such as F2LP, where time has to be
discretized. While other approaches such as ASPMT do support
continuous domains, their reliance on SMT solvers makes 
constraints and dependencies among
variables be lost 
during grounding. The use of s(CASP) brings other advantages: for
example, in a query-driven system the
trace of the proof / generation of the model is a justification for
the answers to a query.  Likewise, explanations for observations via 
abduction are also generated for free, thanks to the goal-directed,
top-down execution of s(CASP).

To the best of the authors' knowledge, our approach is the only one
that faithfully models continuous-time Event Calculus under the stable model
semantics. All other approaches discretize time and thus do not model
EC in a sound manner. Our approach supports both
deduction and abduction with little or no additional effort.

The work reported in this paper can be seen as one of the first non-trivial
applications of s(CASP). It illustrates the
advantages that goal-directed ASP systems have over grounding and SAT
solver-based ones for certain domains / classes of applications.
The Event Calculus and its realization through s(CASP) is being used
to model real-world avionics systems in the aerospace
industry~\cite{hcvs21}. Avionics systems are cyber-physical system 
consisting of sensors and actuators. Sensors are modeled as fluents
while actuator actions correspond to events. The goal is to use
the s(CASP) EC model to verify (timed) properties of these avionics
systems as well as to identify gaps with respect to system
requirements. Abductive capabilities of s(CASP) are being used to
troubleshoot the system as well as to find gaps in system
design. %
The overarching goal of the project is to build 
event-calculus based tools for avionics software systems assurance.


\paragraph {\bf Competing interests:} The author(s) declare none.

\vspace{-.5em}

\renewcommand{\UrlFont}{\small}
\bibliographystyle{acmtrans}

\end{document}